\documentclass[12pt]{article}
\pdfoutput=1

\usepackage{arydshln}
\usepackage[T1]{fontenc}
\usepackage{color}
\usepackage{url}
\usepackage[flushleft]{threeparttable}
\usepackage[utf8]{inputenc}
\usepackage{subfig}
\usepackage[authoryear]{natbib}
   
\usepackage{fullpage}

\usepackage{amsthm}
\theoremstyle{plain}

\usepackage{amsmath}
\usepackage{amssymb}
\usepackage{latexsym}
\usepackage[margin=1.35in]{geometry}
\usepackage{mathabx}
\usepackage{algorithm}
\usepackage{algorithmic}
\usepackage{changepage}

\usepackage{boxedminipage}

\usepackage{listings}

\usepackage{minitoc}

\usepackage{rotating}

\usepackage[flushleft]{threeparttable}
\usepackage{bbm}
\usepackage{multirow}
\usepackage{hyperref}

\usepackage{ifpdf}

\def \argmin{\mathop{\hbox{\rm arg min}}}

\usepackage{graphicx}

\title{\bf GAM(L)A: An econometric model for interpretable Machine Learning
}
\author{Emmanuel Flachaire\footnote{Aix-Marseille University (Aix-Marseille School of Economics), CNRS \& EHESS},
 Gilles Hacheme\footnotemark[1] \hspace{0.05cm}, 
 Sullivan Hué\footnotemark[1] \hspace{0.05cm}
  and S\'{e}bastien Laurent\footnote{Aix-Marseille University (Aix-Marseille School of Economics), CNRS \& EHESS, Aix-Marseille Graduate School of Management -- IAE, France. The authors acknowledge research support by the French National Research Agency Grants ANR-17-EURE-0020 and ANR-21-CE26-0007-01 (project MLEforRisk) and by the Excellence Initiative of Aix-Marseille University - A*MIDEX.}}

\date{January 31, 2022}

\begin{document}

\ifpdf
\DeclareGraphicsExtensions{.pdf, .jpg, .tif}
\else
\DeclareGraphicsExtensions{.eps, .jpg}
\fi

\maketitle

\begin{abstract}
\noindent 
Despite their high predictive performance, random forest and gradient boosting are often considered as black boxes or uninterpretable models which has raised concerns from practitioners and regulators. As an alternative, we propose in this paper to use partial linear models that are inherently interpretable. Specifically, this article introduces GAM-lasso (GAMLA) and GAM-autometrics (GAMA), denoted as GAM(L)A in short. GAM(L)A combines parametric and non-parametric functions to accurately capture linearities and non-linearities prevailing between dependent and explanatory variables, and a variable selection procedure to control for overfitting issues. Estimation relies on a two-step procedure building upon the double residual method. 
We illustrate the predictive performance and interpretability of GAM(L)A on a regression and a classification problem. The results show that GAM(L)A outperforms parametric models augmented by quadratic, cubic and interaction effects. Moreover, the results also suggest that the performance of GAM(L)A  is not significantly different from that of random forest and gradient boosting.

\vspace{2em}
\end{abstract}

\newpage

\section{Introduction}

In recent years, machine learning (ML) algorithms have received considerable attention in the literature and overshadowed traditional econometric models in most applications. Although econometrics and ML have developed in parallel, both approaches allow building predictive models. For that purpose, econometrics relies on probabilistic models describing economic phenomena, whereas ML builds upon smart algorithms learning on their own. However, ML algorithms have recently been shown to be more effective than traditional econometric approaches for modelling complex relationships \citep{Varian2014,Lessmann2015,Charpentier2018,Gunnarsson2021}. Indeed, unlike traditional econometric models, these algorithms are able to capture many complex non-linear relationships through non-parametric approaches, resulting in higher predictive performance. The dominance of ML models in terms of predictive performance, in addition to several other advantages, has led these techniques to be used in several industries. For example, banks and fintech firms are currently considering ML algorithms as challenger models \citep{ACPR2020} in the context of credit scoring, and in some cases ML models are even used for credit production \citep{Hurlin2019}.

However, ML algorithms raise a very important issue for the industry due to their lack of interpretability. Indeed, most of these algorithms are generally considered to be ``black-boxes'', i.e., the opacity of ML techniques leads users to predictions and decision processes that cannot be easily interpreted. The lack of interpretability is currently one of the main limitations of ML algorithms and raises concerns in many applications such as medicine, law, military or finance. ML algorithms need to be interpretable to justify predictions made by the models. For example, in the financial industry, executives need to be able to understand the model to justify their decisions, and regulators require interpretability to ensure fairness of the algorithms.\footnote{See \citet{Barocas2018} for more details on the fairness of ML techniques in a general context and \citet{Hurlin2021} and \citet{Kozodoi2021} in the context of credit scoring.} Furthermore, the lack of interpretability of ML algorithms is currently one of the major concerns of financial regulators regarding the governance of artificial intelligence approaches in the financial industry \Citep{Bracke2019,ACPR2020,EBA2020,EC2020}.

To address this issue, the literature has recently focused on interpretable ML methods.\footnote{Interpretable ML methods seek to explain the behaviour and predictions of ML algorithms. See \Citet{Molnar2020} for more details on interpretable ML.} Specifically, many model-agnostic methods have been proposed to interpret the ex post predictions of black-box models. For example, we can cite here the partial dependence plot \Citep{Friedman2001}, accumulated local effect \Citep{Apley2020}, local interpretable model-agnostic explanations \citep{Ribeiro2016} or SHAP \citep{Lundberg2017}.\footnote{See \citet{Molnar2019} for an overview of interpretable ML methods.} However, interpretations obtained from these methods can be inaccurate representations of the original relationships and potentially mislead users to accept incorrect recommendations, which can be harmful in a high-stakes decision-making context \Citep{Rudin2019}.

Within this context, we rely in this paper on a class of interpretable models. Instead of developing methods to explain the predictions of black boxes, we design flexible models that are fundamentally interpretable. Denoted as GAM-lasso (GAMLA) and GAM-autometrics (GAMA), or GAM(L)A in short, these models combine the predictive performance of ML approaches with the inherent interpretability of econometric models. Formally, this class of models is based on a generalized additive model (GAM) augmented by variables assumed to have a linear effect on the dependent variable. More specifically we consider  interactions of covariate couples. However, due to the possibly large number of interaction variables, we perform variable selection on interactions to avoid overfitting issues. For that purpose, we rely on the lasso \Citep{Tibshirani1996} and autometrics \citep{doornik2009autometrics} approaches. Finally, as the models involve linear (interaction effects) and non-linear (smooth functions of GAM) terms, the variable selection is not performed on raw data but on filtered data using the double residual approach of \citet{Robinson1988}.

Our approach has several advantages. First,  our models are fundamentally interpretable. Indeed, GAM(L)A inherits the simplicity of interpretation of traditional econometric models. Specifically, while smooth functions allow a simple interpretation of the estimated relationships prevailing between dependent and predictive variables, interaction effects can be interpreted as in a simple linear model because they are introduced linearly. Moreover, the effect of the predictive variables can easily be measured from their marginal effects, as in standard econometric models. GAM(L)A thus allows a simple interpretation of prediction and decision processes, unlike ML algorithms. This class of models is also consistent with the recent literature promoting inherently interpretable models instead of interpretable ML methods \Citep{Rudin2019,Rudin2019b,Rudin2021}. 

Second,  as shown in our empirical applications, GAM(L)A competes with sophisticated ML algorithms in terms of predictive performance reinforcing the idea  that  parametric models can have outstanding forecasting performances if they are well specified. 
We also show that ad hoc choices of parametric functional forms aimed at capturing non-linearities, such as quadratic or cubic functions, are too restrictive and may lead to inferior forecasting performances.

We provide a set of Monte Carlo experiments to assess the good performance of GAM(L)A in terms of predictive performance but also its ability to correctly retain the relevant interaction variables. The results suggest that GAM(L)A accurately captures non-linear relationships. Moreover, we show that the double residual approach of \Citet{Robinson1988} leads GAMA (unlike GAMLA) to retain a large number of relevant variables (i.e., high potency) while controlling for the number of retained irrelevant variables (i.e., gauge close to the chosen target size).

Finally, we illustrate the practical usefulness of GAM(L)A using data on regression and classification problems. To that end, we measure the predictive performance of our models using popular measures of performance as well as inference procedures and compare them to the benchmark models in the econometrics and ML literatures. We show that GAM(L)A achieves higher predictive performance than linear models, even when augmented by quadratic and cubic functions. Moreover, the results suggest that our models compete with sophisticated ML algorithms in terms of predictive performance. We also illustrate the interpretability of our new class of models. For that purpose, we assess the parsimony of GAM(L)A and show the simple interpretation of estimated relationships and decision rules through graphical representations of the estimated non-linearities as well as marginal effects. The results suggest that, unlike ML algorithms, GAM(L)A remains interpretable despite capturing complex non-linear relationships.

The remainder of the article is structured as follows. In Section 2, we present the main advantages of ML algorithms as well as the main limitations of traditional linear models and introduce our new class of interpretable models. Section 3 is devoted to Monte Carlo experiments. In Section 4, two empirical applications are proposed to illustrate the potential of GAM(L)A. Finally, we conclude the article in Section 5. Additional figures and tables are provided in Appendix \ref{AppendixA} while Appendix \ref{AppendixB} presents the penalized logistic tree regression of \Citet{Dumitrescu2022}.


\section{Competing with Black Boxes: An Interpretable Parametric Model}

In this section, we present a class of partial linear models that is able to compete with sophisticated ML algorithms in terms of predictive performance while remaining interpretable. The first part of the section describes why ML algorithms achieve high predictive performance and provides a brief presentation of current benchmark ML algorithms. The second part presents the pitfalls of parametric models in that respect. Finally, the last part of the section is devoted to our model.

\subsection{Machine Learning, Non-linear Effects and Interactions}\label{ML}

Consider a regression problem involving a dependent variable $y \in \mathbb{R}$ and a p-dimensional vector of predictive variables $X = \left(X_1, \dots, X_p\right) \in \mathbb{R}^p$. Recent years have witnessed a paradigm shift in terms of the models used to make predictions. Indeed, ML algorithms have progressively replaced parametric models in many applications, when the main objective is to build accurate predictions.
In general, ML algorithms can be defined  as follows:
\begin{equation}
y = f\left(X\right) + \epsilon,
\end{equation}
where $f\left(.\right)$ is a non-parametric function and $\epsilon$ an error term.
The main reason for the growing interest in  ML methods is that these algorithms lead to very high predictive performances and currently outperform parametric models traditionally used by practitioners. The hegemony of ML algorithms over parametric models has been highlighted in several contexts, such as in the credit scoring literature \citep{Paleologo2010, Finlay2011, Lessmann2015}.\footnote{Credit scoring is one of the first fields to which ML algorithms were applied in economics. See, for instance, \citet{Makowski1985}, \citet{Henley1996}, \citet{Desai1996}, and \citet{Baesens2003}.}
The high performance of ML algorithms comes from the fact that the non-parametric functions $f\left(.\right)$ used by these approaches are able to accurately capture complex non-linear effects of covariates including interaction effects. Specifically, instead of specifying a certain relationship between the dependent and explanatory variables, these non-parametric functions $f\left(.\right)$ rely almost exclusively on data to detect non-linearities and interaction effects prevailing between $y$ and $X$. 

Among the many algorithms proposed in the ML literature, ensemble methods such as the random forest \citep{Breiman2001} and gradient boosting \citep{Friedman2001} have been shown to lead to very accurate predictions and even become benchmark models in terms of predictive performance \citep{Lessmann2015,Grennepois2018,Gunnarsson2021}. Random forest and gradient boosting are particular applications of bagging and boosting procedures, which are based on decision trees. The decision tree algorithm is based on a recursive partition of the initial data into smaller homogeneous subsets, in the sense of the dependent variable. Specifically, decision trees recursively split the covariate space into two homogeneous partitions, called nodes, until obtaining nodes that are as homogeneous as possible, which are called terminal nodes or leaves.\footnote{The binary partition corresponds to the CART algorithm \Citep{Breiman1984}, which is the most popular decision tree algorithm.} To do so, for each partition, the algorithm chooses  the most discriminant explanatory variable, i.e., the variable partitioning the original node into the two most homogeneous nodes possible. Formally, a decision tree is defined as
\begin{equation}
	f_{Dt}\left(X\right) = \sum_{m=1}^{M} c_m \mathbb{I}\left(X\in R_m\right),
\end{equation}
where $M$ is the total number of leaves of the tree, $R_m$ is a leaf of the tree and $c_m$ corresponds to the average of the observations' dependent variable in $R_m$.\footnote{More precisely, this equation defines a decision tree in a context of regression, i.e., a regression tree. In the context of classification, $c_m$ corresponds to the dominant class of observations in the leaf $m$.} Despite capturing threshold effects through multiple splits, the predictive performances of decision trees is lacklustre and barely better than random guessing due to their high variance. To solve this issue and achieve higher performance, ensemble methods such as random forest and gradient boosting combine several decision trees.\footnote{By doing so, these algorithms become strong learners, i.e., models that perform substantially better than random predictions, in comparison with weak learners, i.e., models that perform slightly better than random predictions.} 

The random forest algorithm is based on the combination of several decision trees fitted on copies of the original data obtained from a bootstrap procedure.\footnote{In the random forest algorithm, both observations and explanatory variables are randomly selected, whereas in the general bagging procedure, only the observations are.} Denoting by $B$ the total number of trees in the forest, which also corresponds to the number of bootstrap samples, a random forest is defined as 
\begin{equation}
	f_{Rf}\left(X\right) = \frac{1}{B}\sum_{b=1}^{B}f^b_{Dt}\left(X\right),
	\label{RF_eq}
\end{equation}
where $f^b_{Dt}\left(X\right)$ is a decision tree fitted on the $b^{th}$ bootstrap sample. Predictions of the random forest thus simply correspond to averages of individual decision tree predictions. 

Gradient boosting does not involve a bootstrap sampling strategy, but instead, each tree is built upon the errors of the previous decision tree. Formally, gradient boosting is defined as 
\begin{equation}
	f_{Gb}\left(X\right) = \sum_{b=1}^{B}\lambda_b f^b_{Dt}\left(X\right),
	\label{XGB_eq}
\end{equation}
where $B$ is again the total number of trees , which also corresponds to the number of iterations of the algorithm, while $\lambda_b$ represents the weight attributed to the $b^{th}$ tree. From the aggregation of several decision trees, random forest and gradient boosting capture potentially many complex non-linearities, reason why they perform so well.

\subsection{Pitfalls of Parametric Models}\label{Linear Models and non-linearity}

Historically, the simplest approach used to predict $y$ is to rely on the following parametric model
\begin{equation}
	y = X \beta + \epsilon,
	\label{LinearModel}
\end{equation}
where $\beta$ is the vector of parameters to estimate. This simple model assumes a linear relationship between the dependent variable and the predictive variables. However, in practice, the true relationship prevailing between $y$ and $X$ may not be linear but more complex, involving non-linearities and interaction effects. For this reason, the performance of this simple parametric model is lacklustre compared to those of the random forest or gradient boosting algorithms in the presence of complex relationships.
To avoid this pitfall of the simple parametric model, a common approach is to augment Eq.\eqref{LinearModel} with parametric functions of $X$, i.e., 
\begin{equation}
y = X^*\beta + \epsilon,
\label{Parametric_functional_forms}
\end{equation} 
where $X^*$ is a $K$-dimensional vector of predictive variables obtained from transformations of $X$. The objective of $X^*$ is to capture potential non-linearities through parametric transformations of $X$.  Typical examples of parametric functions frequently used to capture non-linearities are the quadratic and cubic functions as well as interactions between covariates,\footnote{It is also possible to include principal components of explanatory variables as well as some of their powers to capture non-linear effects. See, for example, \citet{Castle2010} and \Citet{Castle2013}.} i.e.,
\begin{equation}
X^* = \left(X^1,X^2,X^3,I\right),
\label{X_agumented}
\end{equation} 
where $X^h = \left(X^h_1, \dots, X^h_q\right)$, $I = \left(X_1X_2, \dots, X_{q-1}X_q\right)$ is the set of covariate couples, and $q\leq p$.\footnote{This condition allows excluding meaningless transformations of elements of $X$, like powers of binary variables.}

However, this approach is subject to overfitting issues. Indeed, the model can involve a very large number of parameters to estimate because the number of predictors depends on the number of original explanatory variables $p$. For example, Eq.\eqref{X_agumented} involves $K=75$ predictors for $p=q=10$. Consequently, performing traditional estimation on this high-dimensional model can obviously lead to overfitting. To solve this issue, several approaches have been proposed in the literature to reduce the number of parameters to estimate.

An early approach developed in the literature is lasso. Proposed in the seminal paper of \Citet{Tibshirani1996}, lasso is a penalized regression that proceeds to both estimation and variable selection. To do so, the approach is based on a penalty term that regularizes coefficients and performs variable selection. Considering the linear model associated with Eq.\eqref{Parametric_functional_forms}, lasso solves the following penalized regression problem 
\begin{equation}
	\hat{\beta} = \underset{\beta}{\argmin}\left[\left(y - X^*\beta\right)^T\left(y - X^*\beta\right) + \lambda\sum_{k=1}^{K}|\beta_k|\right].
	\label{Lasso}
\end{equation}
The parameter $\lambda$ is a tuning parameter controlling for the degree of regularization and is generally selected via k-fold cross-validation or information criteria. For cross-validation, two approaches can be considered to select the optimal value of $\lambda$: the value minimizing a loss function applied to the prediction errors (of a training sample), $\hat{\lambda}^{min}$, or the value associated with the most parsimonious model within a 1-standard-error interval, $\hat{\lambda}^{1se}$.\footnote{This method was proposed by \Citet{Breiman1984}.} For comparison purposes, we also consider the adaptive lasso of \citet{Zou2006} which is a popular extension of the lasso method satisfying the oracle property \citep{Fan2001}, unlike the standard lasso estimator.

An alternative approach for variable selection is autometrics \citep{doornik2009autometrics}. This algorithm performs automatic selection model based on the ``Hendry'' general-to-specific model selection \citep{Hendry2000}. Specifically, this approach starts from a generalized unrestricted model (GUM) that includes every potential relationship existing between the dependent variable and predictors, i.e., dynamic effects, breaks, trends, outliers and non-linearities. To reduce the dimension of the GUM, the algorithm performs a battery of tests to eliminate insignificant variables and find a congruent parsimonious model. Rather than testing for each possible sub-model from the GUM, autometrics performs a tree search that reduces the computation time and allows the approach to be feasible even in a high-dimensional context. Finally, the algorithm selects the sub-model encompassing the GUM in the representation of the relationship of concern and passing a battery of diagnostic tests. The diagnostic tests are for instance the error correlation test \citep{Godfrey1978}, the ARCH test \Citep{Engle1982}, the normality test \Citep{Doornik2008}, the heteroscedasticity test \Citep{White1980} and the RESET test \Citep{Ramsey1969}.\footnote{Some of these tests are discarded when autometrics is applied on cross-section data.}
One interesting property of autometrics is that the modeler can choose the target size $\alpha$, which corresponds to the expected percentage of irrelevant variables surviving to the reduction procedure. This parameter is a tuning parameter that solely depends on the modeler's leniency regarding irrelevant variables. For example, a liberal choice could be to fix the target size $\alpha = 0.05$, whereas a more conservative user might prefer lower values, such as $\alpha = 0.01$.\footnote{There is currently no theoretical result on the properties of Autometrics, but simulation results can be found in \citet{Hendry2014}. See \citet{Hendry2011}, \citet{Johansen2016} for theoretical results on properties of Autometrics' related algorithms.}

\subsection{A Partial Linear Approach: GAM-lasso and GAM-autometrics}

Although variable selection methods can address overfitting issues, large differences can still be observed between performances of ML algorithms and parametric models augmented with non-linear transformations of the raw series. 
Therefore, if large differences are observed between the forecasting performance of the random forest and gradient boosting on one hand and linear models on the other hand, this implies that the chosen parametric functions fail to capture the true non-linearities. 
Moreover, if non-linearities are not accurately captured by parametric transformations, it may also disrupt the estimation of interaction effects, leading to an inconsistent selection of these interactions.

For that reason, we propose to use the following class of partial linear models:
\begin{equation}
y = Z\gamma + \sum_{j=1}^{p}g_j\left(X_j\right) + \epsilon,
\label{PartialLinearModel}
\end{equation}
where $Z$ is a vector of $S$ explanatory variables entering linearly in the model, $\gamma$ is the vector of parameters associated to $Z$, and $g_j\left(.\right)$ is a non-parametric function. For the sake of simplicity, we consider the particular case where $Z=I$ so that $Z$ contains solely the $S=p(p-1)/2$ interactions of covariate couples of the variables belonging to $X$. However, $Z$ can be extended to contain any variables provided that the condition stated in Eq.\eqref{Condition_GAM(L)A} below is satisfied.
Model \eqref{PartialLinearModel} allows to capture both interaction effects (introduced linearly) and non-linearities of covariates (from non-parametric functions) and is therefore expected to capture complex relationships.

Unlike ML algorithms, Model \eqref{PartialLinearModel} has the  advantage of keeping the model interpretable. 
Indeed, the linearity assumption on the  effect of $I$ implies that marginal effects of Eq.\eqref{PartialLinearModel} can be computed as follows:
\begin{equation}
\frac{\partial y}{\partial X_j} = c_j +  g_j'\left(X_j\right),
\label{MarginalEffects}
\end{equation}
where $c_j = X_{\left(-j\right)} \gamma_{j}$, $X_{\left(-j\right)}$ is the $\left(p-1\right)$-dimensional vector of covariates excluding $X_j$, $\gamma_{j}$ is the set of coefficients associated with the $p-1$ pairs of interactions involving $X_j$, and $g_j'\left(X_j\right)$ is the partial derivative of $g_j\left(X_j\right)$ with respect to $X_j$. This assumption simplifies the interpretation of the model because marginal effects correspond to the marginal effects of covariates taken individually, eventually augmented by a constant corresponding to the sum of the interactions' marginal effects (evaluated at a chosen value of $X_{\left(-j\right)}$, e.g., a given quantile of $X_{\left(-j\right)}$ over the training sample). 

One could argue about the linear introduction of interaction effects in the model and relax this assumption by considering a non-linear representation of interactions. This idea has been investigated by \citet{Chouldechova2015}. This approach allows the researcher to potentially model both covariates and interaction effects non-linearly while controlling for overfitting issues through variable selection. Although interesting in terms of predictive performance, this approach damages the interpretability of the model. The interpretation of interaction effects becomes complicated, but more importantly, the marginal effects can no longer be easily computed as in Eq.\eqref{MarginalEffects}. Indeed, the marginal effects would not correspond to standard marginal effects of covariates augmented by a constant but to a much more complex formula depending on the non-linear relationships identified for interaction effects. The linearity assumption on interaction effects thus represents the price to pay to keep the model interpretable.

To estimate non-linearities of covariates, we rely on the non-parametric functions of the generalized additive model (GAM). 
Introduced by \Citet{Hastie2017}, the GAM allows one to relax the assumption of a linear relationship between $y$ and $X$ and automatically captures non-linear effects through smooth functions such as
\begin{equation}
g_j\left(X_j\right) = \sum_{l = 1}^{d}\theta_{j,l}b_{j,l}\left(X_j\right),
\label{GAM_eq}
\end{equation}
where $b_{j,l}\left(.\right)$ is a basis function and $\theta_j = \left(\theta_{j,1}, \dots, \theta_{j,d}\right)$ are the associated parameters. The GAM allows for the simultaneous estimation of $\gamma$ as well as the $p$ smooth functions $g_j\left(X_j\right)$ using the backfitting algorithm, which is based on the following objective function:
\begin{equation}
\left(y - \left(Z\gamma + \sum_{j=1}^{p}g_j\left(X_j\right)\right)\right)^T\left(y - \left(Z\gamma + \sum_{j=1}^{p}g_j\left(X_j\right)\right)\right) + \psi\sum_{j=1}^{p}\int \left[g_j''\left(t\right)\right]^2dt,
\label{Obj_func_GAM}
\end{equation}
where $\psi$ is a smoothing (or tuning) parameter. Similar to Eq.\eqref{Lasso}, the objective function of the GAM is penalized. However, the goal of this penalization is not to proceed to variable selection but to avoid overfitting of smooth functions. Indeed, the smoothing parameter $\psi$ prevents non-parametric functions from becoming too wiggly, which could lead to the model having low generalizability. In practice, the smoothing parameter is generally obtained by generalized cross-validation, and the larger the value of $\psi$ is, the smoother the functions.

Similarly to Eq.\eqref{Parametric_functional_forms}, the large number $S$ of parameters to estimate in Eq.\eqref{PartialLinearModel} can potentially lead to overfitting issues and thus requires proceeding to variable selection in $Z$. 
However, as our new class of partial linear models includes both linear (interaction effects) and non-linear (smooth functions) terms, variable selection methods cannot be applied in a standard way. Indeed, partial linear models require a specific estimation method because the presence of non-linear terms can potentially disturb the estimation of linear parameters' coefficients, leading to an inconsistent selection of linear terms. For that purpose, we propose to combine variable selection with the double residual methodology of \Citet{Robinson1988}. The double residual approach follows the Frisch–Waugh–Lovell theorem \Citep{Frisch1933,Lovell1963} and allows one to consistently estimate partial linear models. 
The conditional expectation of the partial linear model defined in Eq.\eqref{PartialLinearModel} is given by 
\begin{equation}
	\mathbb{E}\left(y|X\right) =  \mathbb{E}\left(Z|X\right)\gamma + \sum_{j=1}^{p}g_j\left(X_j\right).
\end{equation}
When subtracted from Eq.\eqref{PartialLinearModel}, it leads to the following equation
\begin{equation}
	y - \mathbb{E}\left(y|X\right) = \left[Z-\mathbb{E}\left(Z|X\right)\right]\gamma + \epsilon.
	\label{y_E_DoubleResiduals}
\end{equation}
If $\mathbb{E}\left(y|X\right)$ and $\mathbb{E}\left(Z|X\right)$ are known, then Eq.\eqref{y_E_DoubleResiduals} can be simply estimated by ordinary least squares (OLS). However, as these conditional expectations are unknown and potentially non-linear, we estimate them using GAM models, i.e.,
\begin{eqnarray}
\mathbb{E}\left(y|X\right)&:&	y = \sum_{j=1}^{p}g_j\left(X_j\right) + u_y,
	\label{eps1}\\
\forall s=1\ldots,S, ~\mathbb{E}\left(Z_s|X\right) &:&	Z_s = \sum_{j=1}^{p}g_j\left(X_j\right) + v_{Z_s},
	\label{eps2}
\end{eqnarray}
where $Z_s$ is the interaction of two different covariates and $u_y$ and $v_{Z_s}$ are error terms. Note that for Model \eqref{eps2} to be estimable, we need to impose a condition of no concurvity between $Z_s$ and the $p$ smooth functions $g_j\left(X_j\right)$, i.e., $Z_s$ cannot be written exactly as the sum of the $p$ smooth functions. This condition holds in our case because $Z=I$ and importantly because we rely on univariate GAM smooth functions to estimate $\mathbb{E}\left(Z_s|X\right)$ and not multivariate kernel methods as in \Citet{Li2007}. Specifically, the following condition must hold: 
\begin{equation}
	E\left(v_Z^\top v_Z\right) \mbox{ is positive definite},
	\label{Condition_GAM(L)A}
\end{equation}
where $v_Z = \left(v_{Z_1}, \dots, v_{Z_S}\right)$.

We then propose to apply lasso and autometrics variable selection methods to the following double residuals model
\begin{equation}
	\hat{u}_y = \hat{v}_Z\delta + \epsilon,
	\label{Double_Residual_Eq}
\end{equation}
where $\hat{u}$  and $\hat v_Z$  represent residuals of Eqs.\eqref{eps1} and \eqref{eps2}, respectively. The double residual approach of \citet{Robinson1988} leads to a root-n-consistent estimate of linear terms' parameters $\delta$, allowing us to correctly perform variable selection on $Z$.\footnote{Note that we only proceed to variable selection on $Z$.} Finally, the smooth functions and parameters of the selected interactions are re-estimated with the following GAM model:
\begin{equation}
	y =  Z^*\gamma^* + \sum_{j=1}^{p}g_j\left(X_j\right) + \epsilon, 
\end{equation}
where $Z^*$ is the subset of $S^*\leq S$ variables in $Z$ selected when estimating
\eqref{Double_Residual_Eq} by lasso or autometrics and $\gamma^*$ is the associated vector of parameters. We denote as GAM-lasso (GAMLA) the resulting model when $Z^*$ is obtained by lasso and as GAM-autometrics (GAMA) when it is obtained by autometrics.\footnote{There already exists an R package that allows one to estimate GAM models penalized by lasso \citep{plsmselect}. However, this package does not apply the double residual approach.}

\section{Simulation Study} \label{SimuSection}

We use Monte Carlo simulations to study the potential of our new approach. Specifically, we assess the predictive performance and variable selection properties of GAM(L)A compared to those of benchmark approaches in the literature. For that purpose, the Monte Carlo simulation is performed on $n_r = 1,000$ replications, with each model being trained on a training sample of $n^{in} = 1,000$ observations and evaluated on $n^{out} = 1,000$ out-of-sample observations, so that $n = n_{in} + n_{out}$ is the total number of simulated data for each replication.

\subsection{Simulation Design}

We generate $p = 10$ predictive variables $x_{i,j}$, $j = 1, \dots,p$, $i = 1, \dots, n$, as well as a response variable $y_i$ according to  the following data generating process (DGP) 
\begin{equation}
	y_i = \sum_{j=1}^{p}\sum_{k = j+1}^{p}\gamma_{j,k}x_{i,j}x_{i,k} + \sum_{j = 1}^{p}g_j\left(x_{i,j}\right) + \epsilon_i,
	\label{DGP}
\end{equation}
where $\epsilon_i \overset{i.i.d.}{\sim} \mathcal{N}\left(0,1\right)$ and $g = \left(g_1, \dots, g_p\right)$ are some non-linear smooth functions defined below. Notice that we use capital letters, e.g., $X$, to refer to a set of variables, small letters with one index ($j$), e.g. $x_j$, to refer to the vector of $n$ observations of a single variable ($j$) and small letters with two indices, e.g., $x_{i,j}$, to refer to one observation ($i$) of a single variable ($j$).

To generate variables displaying non-linearities, we first simulate $q = 5$ variables
$x_{i,j} \overset{i.i.d.} {\sim} \mathcal{N}\left(0,1\right)$ ($ \forall j = 1, \dots, q$) and then use these $q$ variables to obtain $q$ relevant non-linear predictive variables 
$g_j\left(x_{j}\right)$
as follows
\begin{equation}
g_j\left(x_{j}\right) = \left\{
\begin{array}{ll}
\sin\left(5x_{j}\right) & \mbox{for } j = 1, \\
5x_{j} - 5x_{j}\mathbb{I}\left(x_{j}>0\right) & \mbox{for } j = 2, \\
LogN\left(x_{j},0.5\right) & \mbox{for } j = 3, \\
e^{x_{j} }& \mbox{for } j = 4, \\
\arctan\left(10x_{j}\right) & \mbox{for } j = 5, \\
0 & \mbox{otherwise,}
\end{array}
\label{Non_linear_functions_g}
\right.
\end{equation}
where $LogN\left(.\right)$ is the probability density function of a log-normal distribution. Notice that $g_j\left(x_{j}\right)=0$ $\forall j>q$ so that only the first five (out of maximum $p=10$) non-linear variables are relevant. 

The above procedure describes how to generate variables $x_{j}$ and $g_j\left(x_{j}\right)$ for $j=1,\ldots,q$. The remaining $p-q=5$  variables $x_{j}$ for $j=q+1,\ldots,p$ are 
simulated as follows:
\begin{equation}
x_{i,j}	= -g_j\left(x_{i,j-q}\right)/x_{i,j-q} + u_{i,j} \mbox{ for } j = q+1, \dots, p,
\label{Correlation_Variables}
\end{equation}
where $u \sim \mathcal{N}\left(0, 0.4\right)$. These $p$ variables $x_{j}$ are used to create the $p(p-1)/2$ interaction variables $x_{j}x_{k}$ entering in \eqref{DGP}.
Doing so, we introduce dependence between $g_j\left(x_{j}\right)$
 $\forall j\leq q$ and the raw variables $x_{i}$ in order to illustrate the usefulness of the double residual approach in the realistic case 
 of dependence between the interaction variables and the non-linearities $g_j\left(x_{j}\right)$.

\begin{figure}
	\centering
	\vspace{-5em}
	\includegraphics[width=1\linewidth]{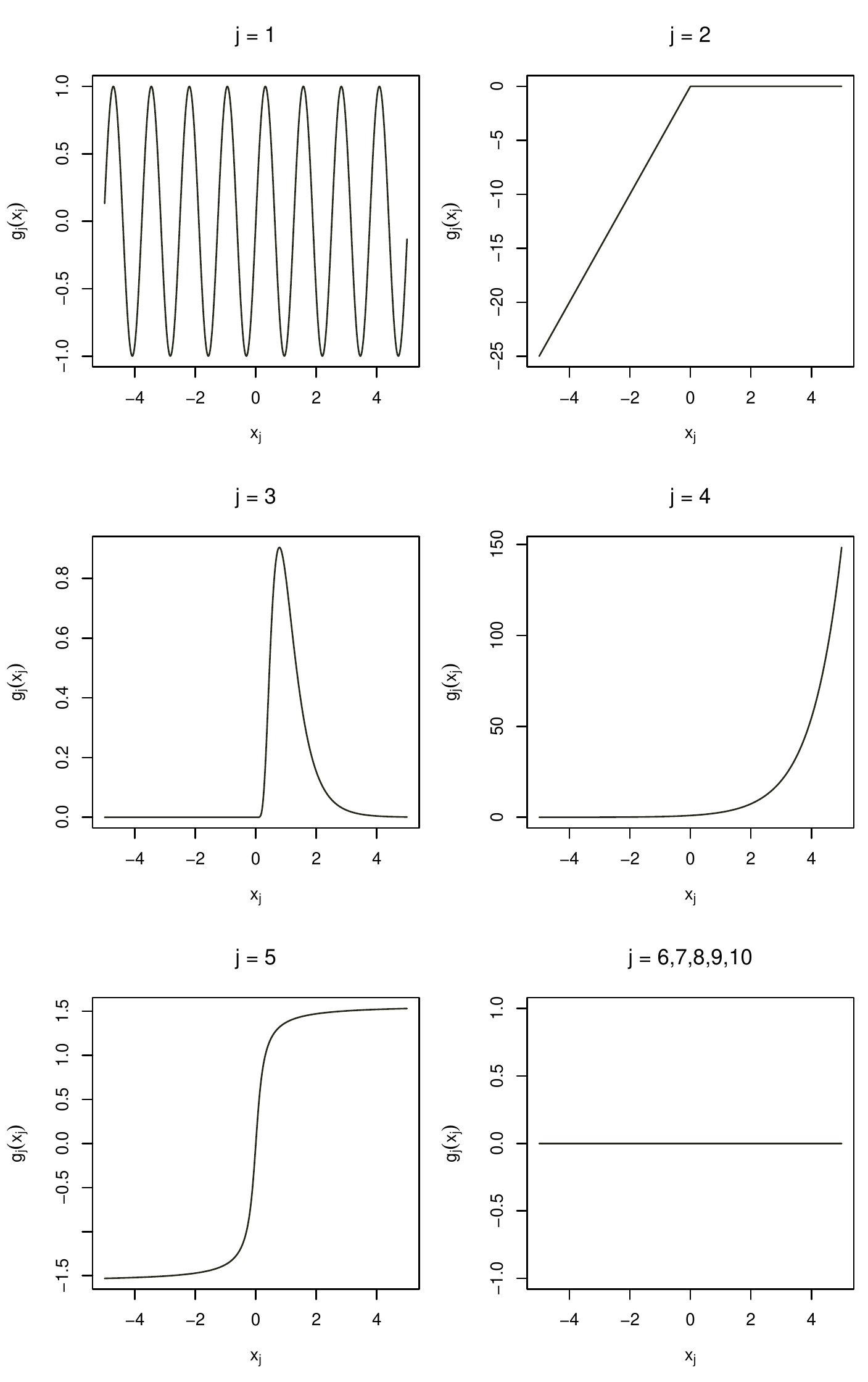}
	\vspace{-0em}
	\caption{Illustrations of non-linear functions $g_j\left(x_{j}\right)$ of Eq.\eqref{Non_linear_functions_g} considered in simulations}
	\label{fig:plot_functions_g_simulations}
\end{figure}

\begin{figure}
	\centering
	\vspace{-5em}
	\includegraphics[width=1\linewidth]{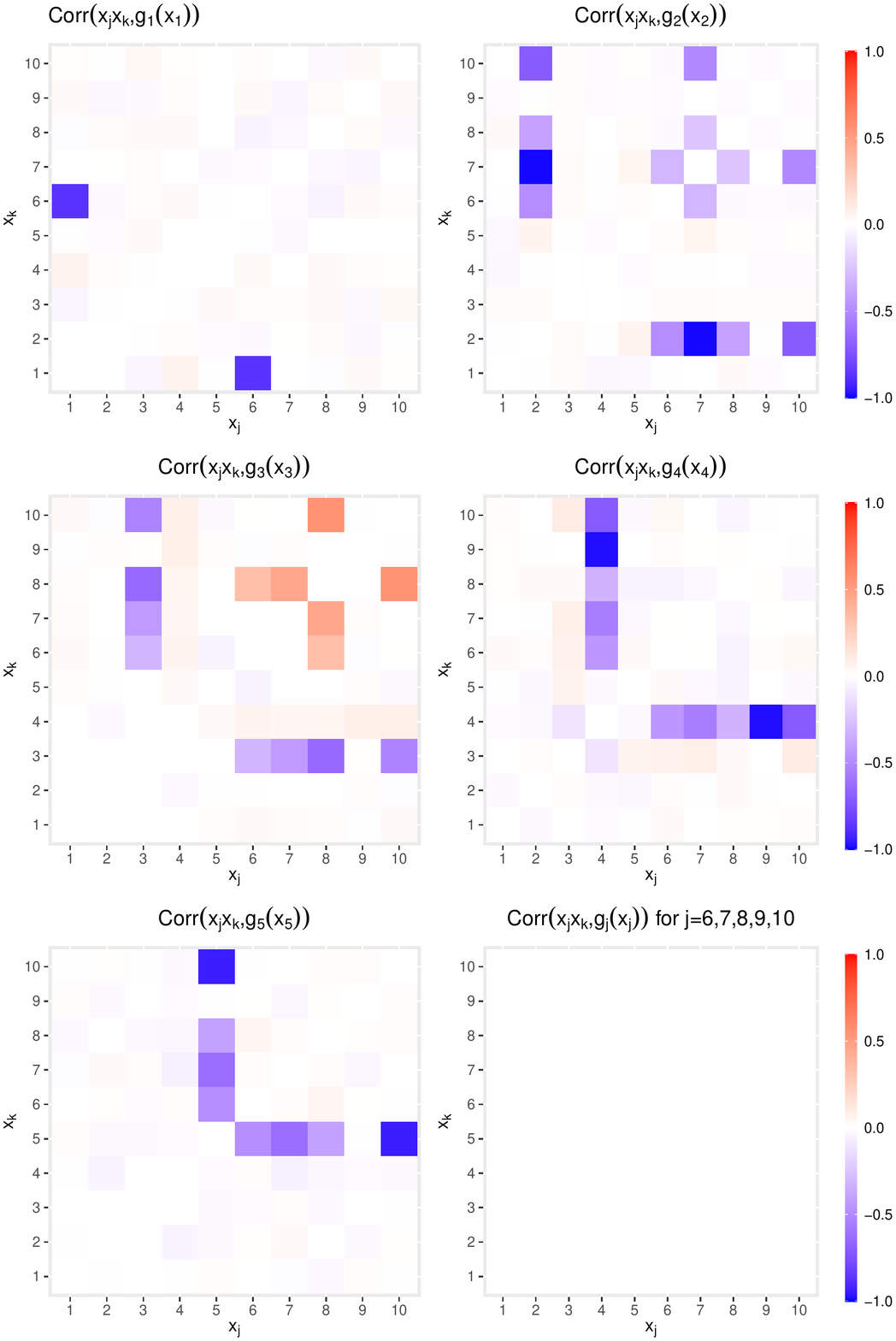}
	\vspace{-0em}
	\caption{Correlation between $x_jx_k$ and $g_j\left(x_{j}\right)$ for one replication of simulations}
	\label{fig:Heatmap}
\end{figure}

The DGP assumes that the first $q$ functions $g$ are non-linear (while the next $p-q$ are redundant) and that some explanatory variables $x_{j}$ (and therefore also interaction variables) are correlated with these non-linear functions. This DGP allows (i) to illustrate the potential of non-parametric functions to accurately capture non-linearities compared to parametric functions and (ii) to highlight the importance of accounting for the correlation between linear and non-linear terms. For the sake of illustration, Figure \ref{fig:plot_functions_g_simulations} plots $g_j\left(x_{j}\right)$ as a function of $x_{j}$ to help visualizing the type of non-linearities introduced in Eq.\eqref{Non_linear_functions_g}, and Figure \ref{fig:Heatmap} displays the correlation between $x_j x_k$ and $g_j\left(x_{j}\right)$ implied by Eq.\eqref{Correlation_Variables}, for one replication of simulations.

To study the finite sample properties of lasso and autometrics in selecting the relevant interaction variables and in rejecting the irrelevant interactions, we set some parameters $\gamma_{j,k}$ to zero and others to a non-zero value. 
Specifically, the relevant interaction effects are $x_{j}x_{j+q}$ for $j = 1,\dots,q$. We set the non-zero coefficients $\gamma_{j,k}$ via a measure of non-centrality  rather than randomly choosing their value. Denoting by $W = \left(x_1x_2, \dots, x_{p-1}x_p, g_1, \dots, g_q\right)$ the matrix of regressors, the coefficients of the relevant variables are defined as
\begin{equation}
\gamma_{j,k} = \xi_{j,k} \sqrt{\mathbb{E}\left[W'W\right]^{-1}_{j,k}},
\end{equation}
where $\xi_{j,k}$ is the non-centrality parameter of interaction 
variable $x_jx_k$.
The non-centrality parameter $\xi_{j,k}$ allows us to calibrate the significance of the $\gamma_{j,k}$ parameters. We consider $\xi_{j,k} = 6$ for $j = 1, \dots, q$, and $k = j+q$ so that these variables have an expected t-statistics (for the null hypothesis that they are redundant) equal to 6. These variables should thus be kept in the final model with a probability close to 1. 

In the simulations, we compare GAM(L)A to several benchmark approaches in the literature.\footnote{For each replication, we consider a cubic basis for non-parametric functions of GAMLA and GAMA.} We consider linear models augmented by quadratic, cubic and interaction terms, as in Eq.\eqref{X_agumented}, whose variables have been selected by lasso, adaptive lasso and autometrics. Denoted as LASSO, A-LASSO and AM, these models allow us to illustrate the failure of parametric functions to accurately capture non-linearities, as well as the inconsistency of variable selection in the presence of both linear and non-linear terms. For each variable selection method, we consider two methods to choose the tuning parameters, i.e.,  $\hat{\lambda}^{min}$ and $\hat{\lambda}^{1se}$ for lasso and $\alpha = 0.05$ and $\alpha = 0.01$ for autometrics. We include in the comparison a naive version of GAMLA and GAMA where the selection of variables in $Z$ is performed by regressing $y$ filtered for the non-linearities in $X$ obtained by GAM (i.e., $\hat{u}_y$) on $Z$ without relying on the double residual approach of \Citet{Robinson1988}, i.e., on $\hat{u}_y = Z\delta + \epsilon$. Denoted as GAMLA* and GAMA*, these models enable us to highlight the absolute need to rely on the double residual approach in the presence of both linear and non-linear terms. We compare GAM(L)A to the methodology of \Citet{Chouldechova2015} which allows us to non-linearly model covariates and interaction effects. Denoted as GAMSEL, this approach is based on lasso to control for overfitting issues. We also compare the predictive performance of GAM(L)A to that of a standard OLS model and that of the current ML benchmarks, i.e., random forest and XGBoost.\footnote{The XGBoost algorithm \citep{Chen2015} is a particular implementation of the gradient boosting algorithm presented in Section \ref{ML} that allows for fast computation and is currently very popular among practitioners and researchers.} Finally, we include in the comparison GAMLA obtained from the adaptive lasso method, denoted as GAMLA (A-LASSO), as well as its version obtained without relying on the double residual approach, denoted as GAMLA* (A-LASSO).

The model performance analysis is conducted using three criteria. First, we study the consistency of variable selection with the potency and the gauge criteria \citep{Castle2011}. Specifically, potency measures the frequency of relevant interaction variables included in the model, such as
\begin{equation}
	\mbox{Potency} = \frac{1}{q}\sum_{j=1}^{q}\mathbb{I}\left(\hat{\gamma}_{j,j+q} \neq 0\right),
\end{equation}
where $\hat{\gamma}_{j,j+q}$ is the coefficient estimated by the model associated with $x_{j}x_{j+q}$, whereas gauge assesses the frequency of irrelevant interactions included in the model, such as
\begin{equation}
	\mbox{Gauge} = \frac{1}{S-q}\sum_{j=1}^{p} \sum_{\begin{subarray}{l} k=j+1, \\ k\neq j+q\end{subarray}}^{p}\mathbb{I}\left(\hat{\gamma}_{j,k} \ne 0\right).
\end{equation}
The optimal variable selection method is thus the one that includes the highest percentage of relevant variables in the model, i.e., the highest potency level, while also controlling for the percentage of irrelevant variables, i.e., the gauge level. Second, we evaluate the predictive performance of models using the mean squared (forecasting) error, which is defined as
\begin{equation}
\mbox{MSE} = \frac{1}{n^{out}}\sum_{i=1}^{n^{out}}\left(\hat{y}_i - y_i\right)^2,
\end{equation}
where $\hat{y}_i$ is the prediction obtained from a model for the $i^{th}$ out-of-sample observation.

\subsection{Evaluation Results}

\begin{table}[!htbp] \centering 
	\caption{Comparison of potency, gauge and MSE under non-linear effects and correlated covariates} 
\resizebox{15cm}{!}{
	\renewcommand{\arraystretch}{1.6}
	\begin{threeparttable}
		\begin{tabular}{lccccc} 
			\hline
			Model & Conditional mean & Tuning parameter  & Potency & Gauge & MSE \\
\hline
			OLS & $(X^1,X^2,X^3,I)\beta$ & & & & $1.478$ \\ 

			LASSO & $(X^1,X^2,X^3,I)\beta$ & $\hat{\lambda}^{min}$ & $0.408$ & $0.212$ & $1.204$ \\ 
			& $(X^1,X^2,X^3,I)\beta$ & $\hat{\lambda}^{1se}$ & $0.018$ & $0.008$ & $1.221$ \\ 
			A-LASSO & $(X^1,X^2,X^3,I)\beta$ & $\hat{\lambda}^{min}$ & $0.392$ & $0.203$ & $1.205$ \\ 
			& $(X^1,X^2,X^3,I)\beta$ & $\hat{\lambda}^{1se}$ & $0.021$ & $0.005$ & $1.221$ \\ 
			AM & $(X^1,X^2,X^3,I)\beta$ & $\alpha = 0.05$ & $0.608$ & $0.058$ & $1.196$ \\ 
			& $(X^1,X^2,X^3,I)\beta$ & $\alpha = 0.01$ & $0.438$ & $0.023$ & $1.182$ \\ 
			GAMLA* & $I\gamma + \sum_{j=1}^{p}g_j\left(X_j\right)$ & $\hat{\lambda}^{min}$ & $0.385$ & $0.081$ & $1.154$ \\ 
			& $I\gamma + \sum_{j=1}^{p}g_j\left(X_j\right)$ & $\hat{\lambda}^{1se}$ & $0.001$ & $0.000$ & $1.194$ \\ 
			GAMLA* (A-LASSO) & $I\gamma + \sum_{j=1}^{p}g_j\left(X_j\right)$ & $\hat{\lambda}^{min}$ & $0.387$ & $0.080$ & $1.154$ \\ 
			& $I\gamma + \sum_{j=1}^{p}g_j\left(X_j\right)$ & $\hat{\lambda}^{1se}$ & $0.000$ & $0.000$ & $1.194$ \\
			GAMA* & $I\gamma + \sum_{j=1}^{p}g_j\left(X_j\right)$ & $\alpha = 0.05$ & $0.518$ & $0.050$ & $1.151$ \\ 
			& $I\gamma + \sum_{j=1}^{p}g_j\left(X_j\right)$ & $\alpha = 0.01$ & $0.453$ & $0.027$ & $1.149$ \\ 
			GAMLA & $I\gamma + \sum_{j=1}^{p}g_j\left(X_j\right)$ & $\hat{\lambda}^{min}$ & $0.980$ & $0.252$ & $1.217$ \\ 
			& $I\gamma + \sum_{j=1}^{p}g_j\left(X_j\right)$ & $\hat{\lambda}^{1se}$ & $0.587$ & $0.004$ & $1.128$ \\ 
			GAMLA (A-LASSO) & $I\gamma + \sum_{j=1}^{p}g_j\left(X_j\right)$ & $\hat{\lambda}^{min}$ & $0.656$ & $0.020$ & $1.126$ \\ 
			& $I\gamma + \sum_{j=1}^{p}g_j\left(X_j\right)$ & $\hat{\lambda}^{1se}$ & $0.387$ & $0.000$ & $1.137$ \\ 
			GAMA & $I\gamma + \sum_{j=1}^{p}g_j\left(X_j\right)$ & $\alpha = 0.05$ & $0.930$ & $0.058$ & $1.160$ \\ 
			& $I\gamma + \sum_{j=1}^{p}g_j\left(X_j\right)$ & $\alpha = 0.01$ & $0.848$ & $0.013$ & $1.127$ \\ 
			GAMSEL & $f_{Gs}\left(X\right)$ & $\hat{\lambda}^{min}$ & $0.864$ & $0.555$ & $1.161$ \\ 
			& $f_{Gs}\left(X\right)$ & $\hat{\lambda}^{1se}$ & $0.295$ & $0.103$ & $1.201$ \\ 
			
			Random Forest & $f_{Rf}\left(X\right)$ & & & & $1.182$ \\ 
			XGBoost & $f_{Gb}\left(X\right)$ & & & & $1.215$ \\ 
\hline			
		\end{tabular} 
		\begin{tablenotes}[para,flushleft]
			\small
			\noindent Note: The results displayed in the last three columns correspond to average values of the criteria over $1,000$ replications. The conditional mean of the competing models/estimation methods is provided in the second column (labeled `Conditional mean').
			Recall that $X^h$ refers to the $h$-th power of the $X$ variables while $I$ is the set of covariate couples.
			Potency and gauge are not reported for OLS, random forest and XGBoost because these models do not perform variable selection. 
		\end{tablenotes}
	\end{threeparttable}
	\label{Simu_Setup1}
	}
\end{table} 

Table \ref{Simu_Setup1} reports the average values of potency, gauge and MSE for seventeen competing models. 
Notice that the selection of variables 
is only performed on fully parametric or partial linear models. Furthermore, while the selection of variables is performed on all variables entering linearly in the model, the gauge and potency reported in Table \ref{Simu_Setup1} is for the interaction variables ($I$) only.

The results first suggest that the linear model with the first three powers of all variables as well as all the interaction variables in the conditional mean (denoted OLS in the table) has by far the highest MSE and therefore suffers from over-fitting problems. 

Furthermore, it appears  that interaction variables are poorly selected by both lasso, adaptive lasso and autometrics when the non-linearities are approximated by quadratic and cubic functions of $X$ (i.e., models denoted LASSO, A-LASSO and AM in the table).
Indeed, these models lead to a low potency due to the poor approximation of non-linearities and sometimes a high gauge. 
More specifically, the highest (resp. smallest) potency is $60.8\%$ (resp. $1.8\%$) while the gauge is between $0.05\%$ and $21.2\%$.
This result implies that relevant interaction variables not included in the model are sometimes triangulated by combinations of other irrelevant variables. 
Regarding gauge levels, the values  are decent although lasso and adaptive lasso lead to a high number of irrelevant interactions being included in the model for the $\hat{\lambda}^{min}$ tuning parameter. 

Random forest and XGBoost give similar forecasting performance, with a small advantage for the former in this simulation. 

Interestingly, GAMLA and GAMA achieve the highest predictive performances. Indeed, the MSEs associated with GAMLA and GAMA are lower than those of all the other competing models, with the smallest MSE being obtained for the GAMA model with $\alpha = 0.01$ as tuning parameter. Therefore, the results of the Monte Carlo simulation show that the partial linear models GAM(L)A compete with sophisticated ML algorithms in terms of performance while leading to consistent identification of relevant and irrelevant variables. Qualitatively similar results are observed for GAMLA estimated by the adaptive lasso method.

The main difference between the GAMLA and GAMA models comes from the gauge level. Both the gauge and potency values are high (resp. low) for GAMLA with the tuning parameter $\hat{\lambda}^{min}$ (resp. $\hat{\lambda}^{1se}$). The highest potency value, $0.980$, comes at the cost of a relatively high gauge level, $25.2\%$, whereas a gauge level close to 0 comes at the cost of the potency decreasing to $58.7\%$. Lasso thus leads to either potency values close to 1 (with $\hat{\lambda}^{min}$) or gauge levels close to 0 (with $\hat{\lambda}^{1se}$). Autometrics however leads to high potency values while controlling for a low gauge level, unlike lasso. Specifically, the gauge level corresponds to the target size $\alpha$ of autometrics. Therefore, an autometrics user can choose the gauge desired while selecting a large percentage of relevant variables. Indeed, while the gauge levels are very close to the target sizes considered, i.e., $\alpha = 0.05$ or $\alpha = 0.01$, the potency levels remain high and are equal to $93\%$ and $84.8\%$. For these reasons, we recommend using GAMA because it allows one to control the target size while also selecting a large percentage of relevant variables, unlike GAMLA. Finally, the weakness of GAMLA compared to the GAMA is also valid for GAMSEL. While leading to a competitive MSE close to those of GAMLA and GAMA, GAMSEL does not allow one to both control the target size and select a large percentage of relevant variables because it is based on a lasso penalization. Note that unlike lasso, the adaptive lasso allows to control the gauge level close to 0. Indeed, gauge values of the GAMLA (A-LASSO) are close to 0 for both values of the tuning parameter $\lambda$, but still at the cost of lower potency values similarly to the lasso.

The results also highlight the importance of the double residual approach of \Citet{Robinson1988}. Indeed, the GAMLA* and GAMA* models select even fewer relevant variables than the lasso and autometrics methods with selection of the first three powers of $X$ and the interaction variables (i.e., LASSO and AM). At best, only half of the relevant interactions are included in these models. In comparison, the highest percentage of relevant variables included in the GAMLA and GAMA models is $98\%$. These results imply that the double residual approach allows to correctly estimate interaction effects, leading to consistent variable selection. Similar results are observed for the GAMLA* (A-LASSO) method.

We also display in Tables \ref{Simu_Setup2}-\ref{Simu_Setup4} in Appendix \ref{AppendixA} the results obtained for three other Monte Carlo simulation setups. Specifically, we consider a DGP in which (i) covariates are correlated but functions $g$ are linear, (ii) functions $g$ are non-linear but covariates are independent, and (iii) functions $g$ are linear and covariates are independent. For that purpose, we generate linear functions $g$ such as $g_j\left(x_{j}\right) = \gamma_j x_{j}$, where $\gamma_j \neq 0$ for $j = 1,\dots,5$ and 0 otherwise, and simulate covariates as $x_{j} \sim \mathcal{N}\left(0,1\right)$ for $j= 1,\dots,p$ in the case of independence. The results obtained are consistent with those in Table \ref{Simu_Setup1}: linear models augmented by parametric functions perform as well as GAMLA and GAMA when the relationship prevailing between $y_i$ and $x_i$ is purely linear, while GAMLA* and GAMA* perform similarly to GAMLA and GAMA when linear and non-linear terms are not correlated. Moreover, results displayed in Table \ref{Simu_Setup4} confirm the conclusion of \citet{Epprecht2021} that the adaptive lasso method leads to better variable selection than autometrics when covariates are linear and independent. In this case, both methods lead to very high potency levels, but the adaptive lasso selects slightly less irrelevant variables than autometrics. However, we find that autometric outperforms the adaptive lasso in the other three setups.

\section{Empirical Application}

In this section, we consider regression and classification problems to assess the potential of  GAM(L)A. 

\subsection{Regression Problem: Boston housing market}

First, we illustrate the practical usefulness of the GAM(L)A model based on a regression problem. For that purpose, we use the popular Boston housing market \citep{harrison1978hedonic},  which has already been considered in several contributions to the literature \citep{Belsley1980,Castle2021,Michelucci2021} as well as for the Kaggle competition ``Boston Housing''. Built by the U.S. Census Bureau, this dataset includes 506 instances and 13 explanatory variables on housing prices in the Boston area, two of which are qualitative. See Table \ref{tab:dicboston} in Appendix \ref{AppendixA} for a description of the variables included in the analysis.

We compare the $[1]$ GAMLA, $[2]$ GAMA models to several benchmarks in the literature.\footnote{Similar to the Monte Carlo simulation, we consider a cubic basis for non-parametric functions of GAMLA and GAMA.}$^,$\footnote{We do not include in the comparison the GAMLA*, GAMLA* (A-LASSO) and GAMA* methods because we demonstrated in Section \ref{SimuSection} that these approaches lead to inconsistent variable selection in the presence of non-linear relationships prevailing between the dependent and predictive variables.} We compare the GAM(L)A to the $[3]$ GAMLA obtained from the adaptive lasso estimator, denoted as GAMLA  (A-LASSO).  We consider three parametric models: $[4]$ a simple linear regression model including only linear terms, $[5]$ a linear regression augmented by quadratic and cubic terms, and $[6]$ a linear model including linear, quadratic, cubic and interaction terms. We also consider $[7]$ the GAM to compare the potential of parametric and non-parametric functions to capture non-linearities of covariates. We include in the comparison $[8]$ lasso, $[9]$ adaptive lasso, and $[10]$ autometrics models including linear, quadratic, cubic and interaction terms. We also implement $[11]$ random forest and $[12]$ XGBoost algorithms and consider these models as benchmarks to evaluate the predictive performance of the other approaches. Finally, we compare the GAM(L)A model to the $[13]$ penalized logistic tree regression model (PLTR) of \citet{Dumitrescu2022}.\footnote{Initially proposed for credit scoring applications, the PLTR can easily be adapted to regression problems.} Like GAM(L)A, the PLTR is also intended to improve the predictive performance of traditional linear models by automatically capturing non-linearities. Specifically, it improves the predictive performance of linear models by including univariate and bivariate threshold effects obtained from short-depth decision trees and is penalized to control for the number of threshold effects included in the model. Formally, the PLTR models the conditional mean as $X\beta + \mathcal{V}_1 \xi + \mathcal{V}_2\zeta$, where $\mathcal{V}_1$ and $\mathcal{V}_2$ denote respectively univariate and bivariate threshold effects obtained from decision trees, and $\xi$ and $\zeta$ are the associated parameters to estimate. However, GAM(L)A and PLTR differ substantially in how non-linearities are captured. While GAM(L)A is based on GAM non-parametric functions and interaction effects, the PLTR relies on univariate and bivariate threshold effects obtained from short-depth decision trees. See Appendix \ref{AppendixB} for more details on the PLTR model of \citet{Dumitrescu2022}.

To evaluate the performance of these models, we use a 10-fold cross-validation approach based on the mean squared error (MSE), which is the benchmark performance measure for regression problems. For that purpose, we randomly divide the initial sample into 10 sub-samples of equal size and iteratively consider one sub-sample for prediction, while the nine other sub-samples are used to fit models. The MSE is then computed on the vector including all predictions obtained from each sub-sample. Moreover, we use the model confidence set (MCS) of \citet{Hansen2011} to identify models exhibiting significantly better predictions. Indeed, the MCS identifies the bucket of models that exhibit similar performance and are superior to the remaining models. To do so, we apply the MCS on the vector of squared errors obtained for the 10-folds. 
The MCS is computed from 10,000 bootstrap samples and since we have cross-section data, we chose a block size of one observation. 

Finally, we analyse the interpretability of GAM(L)A. 
However, measuring the interpretability of a model is difficult, and there is currently no consensus on the real definition of an ``interpretable model'' \citep{Molnar2019}. For that reason, we consider two criteria to measure the interpretability of GAM(L)A that correspond to the function and human-level evaluation of interpretability proposed by \citet{Doshi2017}.\footnote{See \citet{Doshi2017} for a detailed description of the interpretability evaluation.} On the one hand, we assess the parsimony of the models and compute the number of interaction effects selected based on the same vector including all predictions used to compute the MSE. The idea of this quantitative criterion is that the fewer the number of variables involved in a prediction, the easier it is for a user to understand the determinants of predictions. On the other hand, we represent smooth functions and marginal effects of GAMA (those of GAMLA are not reported to save space). 

\begin{table}[htbp] 
	\centering 
	\caption{Number of variables selected, MSE and MCS: Boston housing dataset} 
	\footnotesize
	\renewcommand{\arraystretch}{1.6}
	\begin{adjustwidth}{-1cm}{}
		\begin{threeparttable}
			\begin{tabular}{llccccr}
				\hline
				\# & Model & Conditional  & Tuning  & Number of & MSE & MCS P-value \\  
				& & mean & parameter & interactions &  &  \\ 
				\hline
				$[1]$ & GAMLA & $I\gamma + \sum_{j=1}^{p}g_j\left(X_j\right)$ & $\hat{\lambda}^{min}$ & 50 & $10.235$ & $0.920$ \\ 
				&GAMLA & $I\gamma + \sum_{j=1}^{p}g_j\left(X_j\right)$ & $\hat{\lambda}^{1se}$ & 22 & $9.594$ & $1.000$ \\ 
				$[2]$ &GAMA & $I\gamma + \sum_{j=1}^{p}g_j\left(X_j\right)$ & $\alpha = 0.05$ & 27 & $10.086$ & $0.933$ \\ 
				&GAMA & $I\gamma + \sum_{j=1}^{p}g_j\left(X_j\right)$ & $\alpha = 0.01$ & 21 & $10.389$ & $0.920$ \\  
				\hline
				$[3]$ & GAMLA (A-LASSO) & $I\gamma + \sum_{j=1}^{p}g_j\left(X_j\right)$ & $\hat{\lambda}^{min}$ & 19 & $9.974$ & $0.933$ \\ 
				&GAMLA  (A-LASSO) & $I\gamma + \sum_{j=1}^{p}g_j\left(X_j\right)$ & $\hat{\lambda}^{1se}$ & 9 & $10.800$ & $0.720$ \\ 
				\hline
				$[4]$ &OLS & $X\beta$ & & & $23.938$ & $<0.001$ \\ 
				$[5]$ &OLS & $(X^1,X^2,X^3)\beta$ & & & $16.039$ & $0.006$ \\ 
				$[6]$ &OLS & $(X^1,X^2,X^3,I)\beta$ & & 78& $24.079$ & $0.006$ \\ 
				\hline
				$[7]$ &GAM & $\sum_{j=1}^{p}g_j\left(X_j\right)$ & & & $13.186$ & $0.008$ \\ 
				\hline
				$[8]$ &LASSO & $(X^1,X^2,X^3,I)\beta$ & $\hat{\lambda}^{min}$ & 69 & $14.698$ & $0.119$ \\ 
				&LASSO & $(X^1,X^2,X^3,I)\beta$ & $\hat{\lambda}^{1se}$ & 31 & $15.683$ & $0.006$ \\ 
				$[9]$ &A-LASSO & $(X^1,X^2,X^3,I)\beta$ & $\hat{\lambda}^{min}$ & 10 & $22.827$ & $0.005$ \\ 
				&A-LASSO & $(X^1,X^2,X^3,I)\beta$ & $\hat{\lambda}^{1se}$ & 6 & $21.874$ & $0.003$ \\ 
				$[10]$ &AM & $(X^1,X^2,X^3,I)\beta$ & $\alpha = 0.05$ & 32 & $14.881$ & $0.007$ \\ 
				&AM & $(X^1,X^2,X^3,I)\beta$ & $\alpha = 0.01$ & 29 & $15.467$ & $0.006$ \\ 
				\hline
				$[11]$ &Random Forest & $f_{Rf}\left(X\right)$ & & & $10.008$ & $0.952$ \\ 
				$[12]$ &XGBoost & $f_{Gb}\left(X\right)$ & & & $ 9.729$ & $0.955$ \\
				\hline
				$[13]$ &PLTR & $X\beta + \mathcal{V}_1\xi+ \mathcal{V}_2\zeta$ & &  & $13.726$ & $0.024$ \\
				\hline
			\end{tabular} 
			\begin{tablenotes}[para,flushleft]
				\small
				\noindent Note: The conditional mean of the competing models/estimation methods is provided in the second column (labeled `Conditional mean').  The results displayed in column MCS correspond to the mean square (forecast) error over the 10-folds while those in the last column are the p-values of the MCS test based on 10,000 bootstrap samples. 
			\end{tablenotes}
		\end{threeparttable}
	\end{adjustwidth}
	\label{Boston_results} 
\end{table}

Table \ref{Boston_results} displays the results obtained for the 10-folds. The results suggest that  parametric linear models $[4$ - $6]$ yield the worst performance of all models. Compared to $[11]$ random forest and $[12]$ XGBoost algorithms, the MSEs of the linear models are between 1.6 to 2.4 times larger, even when augmented by quadratic and cubic effects as well as interactions of covariates couples. Similarly, the performance of the $[8]$ LASSO, $[9]$ A-LASSO and $[10]$ AM models including quadratic and cubic effects as well as interactions are relatively poor compared to that of random forest and XGBoost, despite being better than that of linear models. The results of the A-LASSO model are even worse as it selects a small number variables compared to LASSO and AM models. In line with the findings of \citet{Dumitrescu2022}, our results suggest that these parametric functions of the raw data are not flexible enough to capture the non-linearity of this data. In contrast, non-parametric functions of $[7]$ GAM accurately capture non-linearities, with the MSE of GAM being closer to that of sophisticated ML algorithms than linear models. However, non-parametric functions are insufficient to achieve the performance of these high-performing algorithms, as linear models and GAM are not included in the subset of outperforming models identified by the MCS, with a nominal level of $\alpha_{MCS} = 20\%$, except for one case, with a nominal level of $\alpha_{MCS} = 10\%$. PLTR is not better than GAM and is also rejected from the MCS. This result implies that the univariate and bivariate threshold effects of the PLTR are insufficient to correctly capture all the non-linearities in the data.\footnote{The predictive performance of the PLTR could be improved by including threshold effects obtained from triplets and quadruplets of explanatory variables.}

Interestingly, the combination of non-parametric GAM functions and interaction effects lead 
to very satisfactory results.  Indeed, $[1]$ GAMLA and $[2]$ GAMA have MSEs very close to those of  Random Forest and XGBoost
and belong to the MCS (at the nominal level of 10\%) together with these two ML algorithms. The MSE of GAMLA with the $\hat{\lambda}^{1se}$ tuning parameter is actually the smallest of all competing models. The results obtained for the $[3]$ GAMLA (A-LASSO) are qualitatively similar to those of GAM(L)A.

The take-away message here is that parametric models can compete with sophisticated ML algorithms in terms of predictive performance, as long as the models are well specified. Therefore, it is not essential to use black boxes to reach high predictive performance: parametric models can fulfill the same objective. 

\citet{Castle2021} investigate the same dataset and find it essential to model interaction effects between explanatory variables and a dummy variable distinguishing Boston and its suburbs.\footnote{\citet{Belsley1980} note that observations 357 to 488 correspond to Boston, while other observations correspond to the suburbs.} We display in Table \ref{Boston_results_dummy} in Appendix \ref{AppendixA} the results obtained from this analysis. We find that including a dummy variable for Boston and its suburbs slightly increases the predictive performance of some models, such as GAMA and PLTR, but slightly decreases the performance of other ones, such as AM and random forest.  However, results are qualitatively similar and the conclusions remain the same.

\begin{figure}
	\centering
	\vspace{-9em}
	\includegraphics[width=1\linewidth]{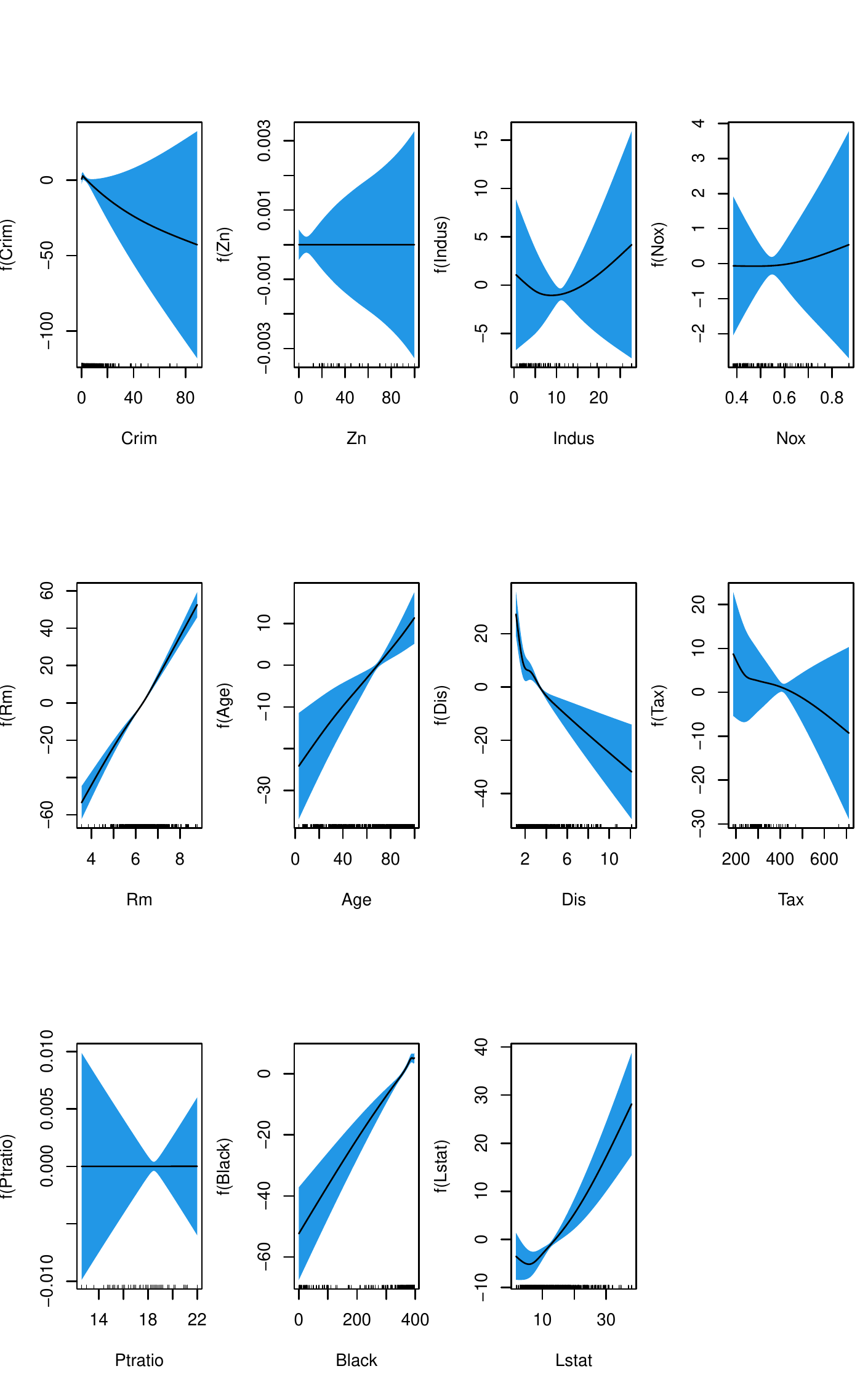}
	\vspace{-1em}
	\caption{Non-parametric functions estimated for GAMA associated with the $\alpha = 0.05$ target size: Boston housing dataset}
	\label{fig:Gam_functions_GAMLA_1se_Boston}
\end{figure}

\begin{figure}
	\centering
	\vspace{-8em}
	\includegraphics[width=1\linewidth]{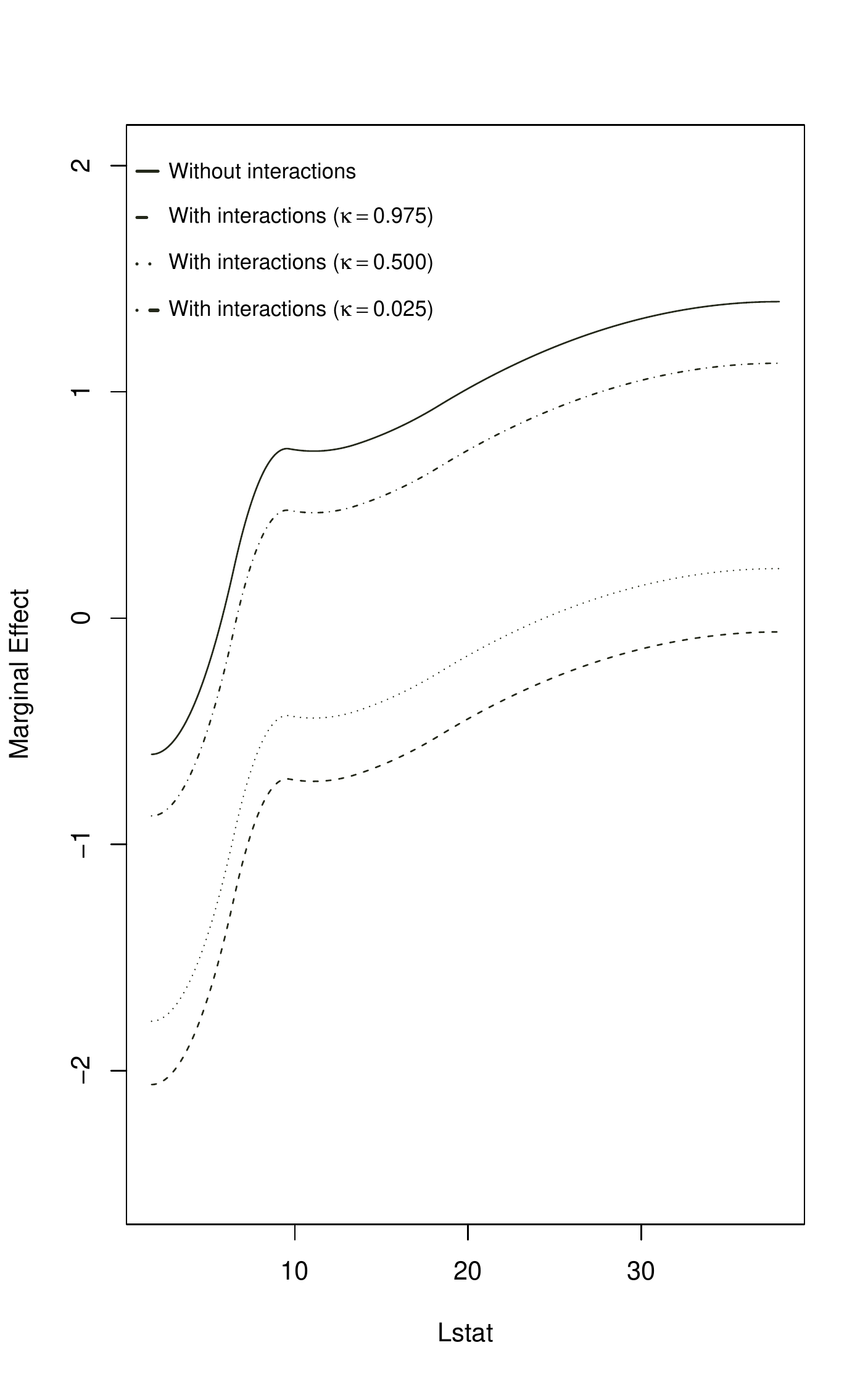}
	\vspace{-2em}
	\caption{Marginal Effect of Lstat (with and without interactions) for GAMA associated with the $\alpha = 0.05$ target size: Boston housing dataset}
	\label{fig:Marginal_Effect_Lstat_Translation_Interactions}
\end{figure}

GAMLA and GAMA are also interpretable, unlike sophisticated ML algorithms. 
For the sake of illustration, Figure \ref{fig:Gam_functions_GAMLA_1se_Boston} displays estimated non-parametric functions of GAMA associated with a target size $\alpha = 0.05$.\footnote{Table \ref{tab:Boston_GAM_results} in Appendix \ref{AppendixA} displays the estimation results of non-parametric functions of GAMA associated with a  target size $\alpha = 0.05$.} The results suggest that while the effect of the variable Rm is almost linear, for some other variables, the effects are highly non-linear, such as
Dis, which exhibits a partial linear effect, or Lstat that displays a quadratic effect. 
As explained above, such graphs are however not available for random forest and XGBoost, which makes these methods difficult to interpret. At best, graphs of the variable importance criterion, as illustrated in Figure \ref{fig:Variable_Importance} in Appendix \ref{AppendixA}, are available for these two algorithms. These graphs represent a measure of predictive power of each variable in the considered algorithm.
Although these graphs allow to rank variables from the least important one to the most important one, they does not say anything about the nature and even the sign of the non-linearities. 

As shown in Eq.\eqref{MarginalEffects} marginal effects are easily computable for GAM(L)A. Figure \ref{fig:Marginal_Effect_Lstat_Translation_Interactions} displays the marginal effect of the variable Lstat obtained for GAMA with a  target size $\alpha = 0.05$. The solid curve represents the marginal effect of Lstat when ignoring all the retained interaction variables involving Lstat, i.e., when $c_j=0$. The graph shows that the marginal effect is non-linear and increases with the level of Lstat. The effect is negative (resp. positive) when Lstat is smaller (resp. larger) than $ 5.87$. However, GAMA selected four variables  interacting with Lstat, i.e., Age, Black, Dis and Tax. Therefore, marginal effects of Lstat should better be evaluated for some representative values of these variables (denoted $X_{\left(-j\right)}$).  The three additional curves plotted in Figure \ref{fig:Marginal_Effect_Lstat_Translation_Interactions} correspond to the marginal effect of Lstat when the elements of $X_{\left(-j\right)}$ are set to their 2.5, 50.0 and 97.5\% quantiles. The effects of the interaction variable being assumed to be linear, the four curves are parallel. These interaction variables therefore imply a change in the level of marginal effects.


\subsection{Classification Problem: Credit scoring}

Second, we consider a credit scoring application. The goal of credit scoring is to predict customer default and is based on the estimation of customers' default probability. To this end, we use the ``Credit Card'' dataset \citep{greene2003econometric} that has also been used for Kaggle competitions. The dataset includes 1,319 observations, and the dependent variable corresponds to the acceptation and rejection of customers' credit card applications. To explain the application decision, we can rely on   11 explanatory variables available in the database, two of which being qualitative, and one continuous variable taking only two values. See Table \ref{tab:cred} in Appendix \ref{AppendixA} for a description of the variables.

We compare the same models as those previously considered in the Boston housing application. However, $[1]$ GAMLA, $[2]$ GAMA, $[3]$ GAMLA (A-LASSO), $[4$ - $6]$ OLS, $[7]$ GAM, $[8]$ LASSO, $[9]$ A-LASSO and $[10]$ AM are linear probability models instead of traditional linear models because the dependent variable is binary.\footnote{We rely on linear probability models instead of general linear models such as logistic and probit regressions for two reasons. First, several authors highlighted advantages of linear probability models over logistic and probit regressions. See \citet{Angrist2008},\citet{Wooldridge2015}, and \citet{Boucher2020}  among others. Second, as the dependent variable of Eq. \eqref{Double_Residual_Eq} is not a binary variable but a residual the double residual methodology cannot be combined used with logistic and probit regressions.}\footnote{The PLTR is estimated by a logit regression as proposed in \Citet{Dumitrescu2022}.} To assess the performance of the models, we rely on a 10-fold cross-validation approach based on the area under the ROC curve (AUC), which is the benchmark performance measure for classification problems.\footnote{As probability linear models can lead to probabilities smaller than $0$ and larger than $1$, we set predicted probabilities between $0$ and $1$ to compute the AUC.} The AUC measures the link between the false and true positive rates over every possible threshold between 0 and 1. Specifically, the AUC represents the probability that the occurrence of a random bad application is higher than that of a random good application. Moreover, to assess whether the AUC difference between two models is significant, we use the pairwise AUC test used in \Citet{Candelon2012}. 

\begin{table}[htbp!] 
	\centering 
	\caption{Number of variables selected and AUC: Credit card dataset} 
	\footnotesize
	\renewcommand{\arraystretch}{1.6}
	\begin{adjustwidth}{0cm}{}
		\begin{threeparttable}
			\begin{tabular}{llcccc}
				\hline
				\# & Model & Conditional & Tuning & Number of & AUC \\ 
				 &  & mean & parameter & interactions &  \\ 
				\hline  
				$[1]$ & GAMLA & $I\gamma + \sum_{j=1}^{p}g_j\left(X_j\right)$ & $\hat{\lambda}^{min}$ & 9 & $0.995$ \\ 
				& GAMLA & $I\gamma + \sum_{j=1}^{p}g_j\left(X_j\right)$ & $\hat{\lambda}^{1se}$ & 0 & $0.995$ \\ 
				$[2]$ & GAMA & $I\gamma + \sum_{j=1}^{p}g_j\left(X_j\right)$ & $\alpha = 0.05$ & 3 & $0.995$ \\ 
				& GAMA & $I\gamma + \sum_{j=1}^{p}g_j\left(X_j\right)$ & $\alpha = 0.01$ & 2 & $0.996$ \\
				\hline
				$[3]$ & GAMLA (A-LASSO) & $I\gamma + \sum_{j=1}^{p}g_j\left(X_j\right)$ & $\hat{\lambda}^{min}$ & 9 & $0.995$ \\ 
				& GAMLA (A-LASSO) & $I\gamma + \sum_{j=1}^{p}g_j\left(X_j\right)$ & $\hat{\lambda}^{1se}$ & 0 & $0.995$ \\ 
				\hline 
				$[4]$ & OLS & $X\beta$ & & & $0.924$ \\ 
				$[5]$ & OLS & $(X^1,X^2,X^3)\beta$ & & & $0.967$ \\ 
				$[6]$ & OLS & $(X^1,X^2,X^3,I)\beta$ & & 55& $0.988$ \\ 
				\hline
				$[7]$ & GAM & $\sum_{j=1}^{p}g_j\left(X_j\right)$ & & & $0.995$ \\ 
				\hline
				$[8]$ & LASSO & $(X^1,X^2,X^3,I)\beta$ & $\hat{\lambda}^{min}$ & 35 & $0.964$ \\ 
				& LASSO & $(X^1,X^2,X^3,I)\beta$ & $\hat{\lambda}^{1se}$ & 27 & $0.963$ \\ 
				$[9]$ & A-LASSO & $(X^1,X^2,X^3,I)\beta$ & $\hat{\lambda}^{min}$ & 10 & $0.968$ \\ 
				& A-LASSO & $(X^1,X^2,X^3,I)\beta$ & $\hat{\lambda}^{1se}$ & 7 & $0.972$ \\
				$[10]$ & AM & $(X^1,X^2,X^3,I)\beta$ & $\alpha = 0.05$ & 9 & $0.982$ \\ 
				& AM & $(X^1,X^2,X^3,I)\beta$ & $\alpha = 0.01$ & 5 & $0.985$ \\ 
				\hline
				$[11]$ & Random Forest & $f_{Rf}\left(X\right)$  & & & $0.995$ \\ 
				$[12]$ & XGBoost & $f_{Gb}\left(X\right)$ & & & $0.996$ \\
				\hline
				$[13]$ & PLTR & $X\beta + \mathcal{V}_1\xi + \mathcal{V}_2\zeta$ & & & $0.996$ \\
				\hline
			\end{tabular} 
			\begin{tablenotes}[para,flushleft]
				\small 
				\noindent Note: The conditional mean of the competing models/estimation methods is provided in the second column (labeled `Conditional mean'). The results in the last column correspond to the AUC computed on the predictions for the 10-folds.
			\end{tablenotes}
		\end{threeparttable}
	\end{adjustwidth}
	\label{CreditCard_results} 
\end{table}


\begin{sidewaystable}[htbp!] 
	\centering 
	\caption{P-values of pairwise bilateral AUC tests: Credit card dataset} 
	\renewcommand{\arraystretch}{1.6}
	\begin{adjustwidth}{0cm}{}
		\begin{threeparttable}
			\begin{tabular}{l|cccccccccc}
				\hline
				& $[1]$ & $[1]$ & $[2]$ & $[2]$ & $[3]$  & $[3]$ & $[7]$ & $[11]$ & $[12]$ & $[13]$ \\
				\hline
				$[1]$ GAMLA ($\hat{\lambda}^{min}$) & - & $0.591$ & $0.269$ & $0.189$ & $0.424$ & $0.591$ & $0.731$ & $0.768$ &$0.666$ & $0.149$ \\
				$[1]$ GAMLA ($\hat{\lambda}^{1se}$) &  & - &    $0.769$   &    $0.485$   &    $0.413$   &   $1$    & $0.753$ & $0.637$ & $0.807$ & $0.241$ \\
				$[2]$ GAMA ($\alpha = 0.05$) &  &  & - &   $0.541$    &   $0.196$    &   $0.769$    & $0.643$ & $0.592$ & $0.856$ & $0.259$ \\
				$[2]$ GAMA ($\alpha = 0.01$) &  &  &  & - &   $0.092$    &   $0.485$    & $0.441$ & $0.508$ & $0.951$ & $0.350$\\
				$[3]$ GAMLA (A-LASSO, $\hat{\lambda}^{min}$) &  &  &  &  & - & $0.413$ & $0.462$ & $0.964$ & $0.446$ & $0.108$\\
				$[3]$ GAMLA (A-LASSO, $\hat{\lambda}^{1se}$) &  &  &  &  &  &  - & $0.753$ & $0.637$ & $0.807$ & $0.241$ \\
				$[7]$ GAM &  & &  &  &  &  & - & $0.679$ & $0.757$ & $0.218$ \\
				$[11]$ Random Forest &  &  &  &  &  &  &  & - & $0.516$ & $0.129$ \\
				$[12]$ XGBoost &  &  &  &  &  &  &  &  & - & $0.406$ \\
				$[13]$ PLTR &  &  &  &  &  &  &  &  & & -\\
				\hline
			\end{tabular} 
			\begin{tablenotes}[para,flushleft]
				\small
				\noindent Note: This table displays the p-values of pairwise bilateral test of AUC \citep{Candelon2012} for the GAMLA, GAMA, GAMLA (A-LASSO), GAM, Random Forest, XGBoost and PLTR models. P-values associated with the OLS, LASSO, A-LASSO and AM models are not reported here but are all smaller than $0.001$ when compared to these models.
			\end{tablenotes}
		\end{threeparttable}
	\end{adjustwidth}
	\label{Pairwise_AUC_Pvals} 
\end{sidewaystable} 

Table \ref{CreditCard_results} displays the number of interaction variables selected as well as the AUC for each of the competing models. Similarly to the Boston housing dataset, $[4$ - $6]$ purely parametric linear models lead to low predictive performance compared to more sophisticated models. 
Indeed, the AUC of the parametric linear models, $[8]$ LASSO, $[9]$ A-LASSO and $[10]$ AM models are much lower than those of $[11]$ random forest and $[12]$ XGBoost meaning that quadratic functions, cubic functions and  interactions of covariate couples are not sufficient to accurately capture non-linearities in this application.

In contrast to linear parametric models, non-parametric $[7]$ GAM functions capture non-linearities much better, judging by their better performance. More precisely, GAM, $[1]$ GAMLA, $[2]$ GAMA and $[3]$ GAMLA (A-LASSO) achieve similar performance to random forest, XGBoost and $[13]$ PLTR. These results highlight the potential of non-parametric GAM functions to capture non-linearities because the PLTR, random forest and XGBoost have been identified as benchmark models for credit scoring applications \citep{Lessmann2015,Grennepois2018,Dumitrescu2022,Gunnarsson2021}. Moreover, the results also suggest that the predictive performance of these models mostly comes from non-linearities of covariates rather than interaction effects. Indeed, unlike the Boston housing application, GAM leads to similar performance to GAMLA, GAMA, random forest and XGBoost, implying that the importance of interaction effects is negligible for this dataset. To confirm this result, we display in Table \ref{Pairwise_AUC_Pvals} the p-values of a pairwise bilateral test of the AUC, which tests whether the difference between the AUCs of two competing models is significant. For the sake of clarity, results related to the OLS, LASSO, A-LASSO and AM models are not displayed, but we find that these models are rejected because their p-values are all inferior to $0.001$.
Regarding GAMLA, GAMA, GAMLA (A-LASSO), GAM, random forest, XGBoost and PLTR, the results show that these models lead to similar AUCs because none of the p-values are lower than the nominal level of $5\%$. This result is also confirmed by the small number of interaction effects selected by the GAMLA, GAMA and  GAMLA (A-LASSO) models, i.e., fewer than $10$ interactions. 

\begin{figure}
	\centering
	\vspace{-9em}
	\includegraphics[width=1\linewidth]{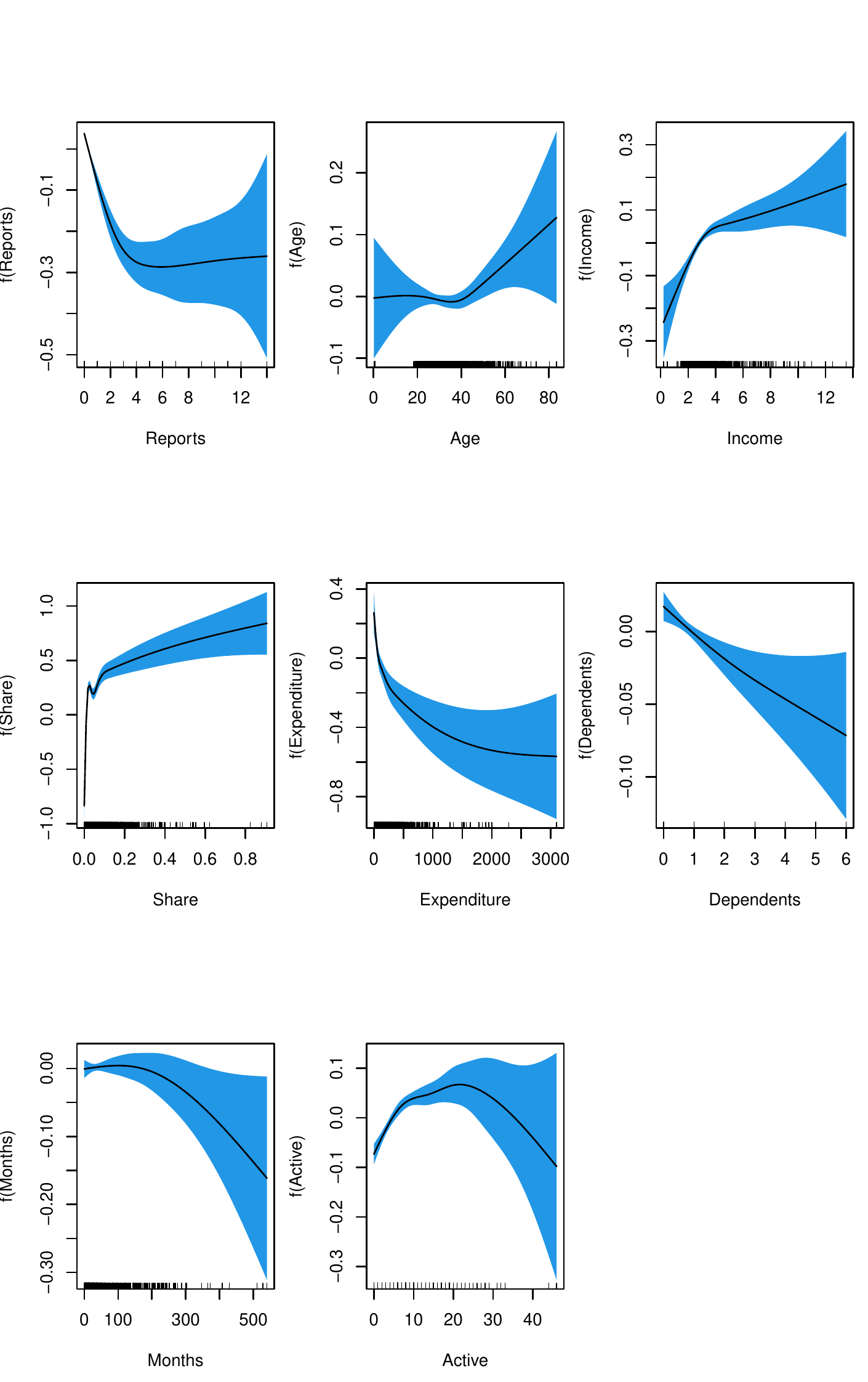}
	\vspace{-1em}
	\caption{Non-parametric functions estimated for GAMA associated with a target size $\alpha = 0.05$ : credit card dataset}
	\label{fig:Gam_functions_GAMA_lib_CreditCard}
\end{figure}

\begin{figure}
	\centering
	\vspace{-8em}
	\includegraphics[width=1\linewidth]{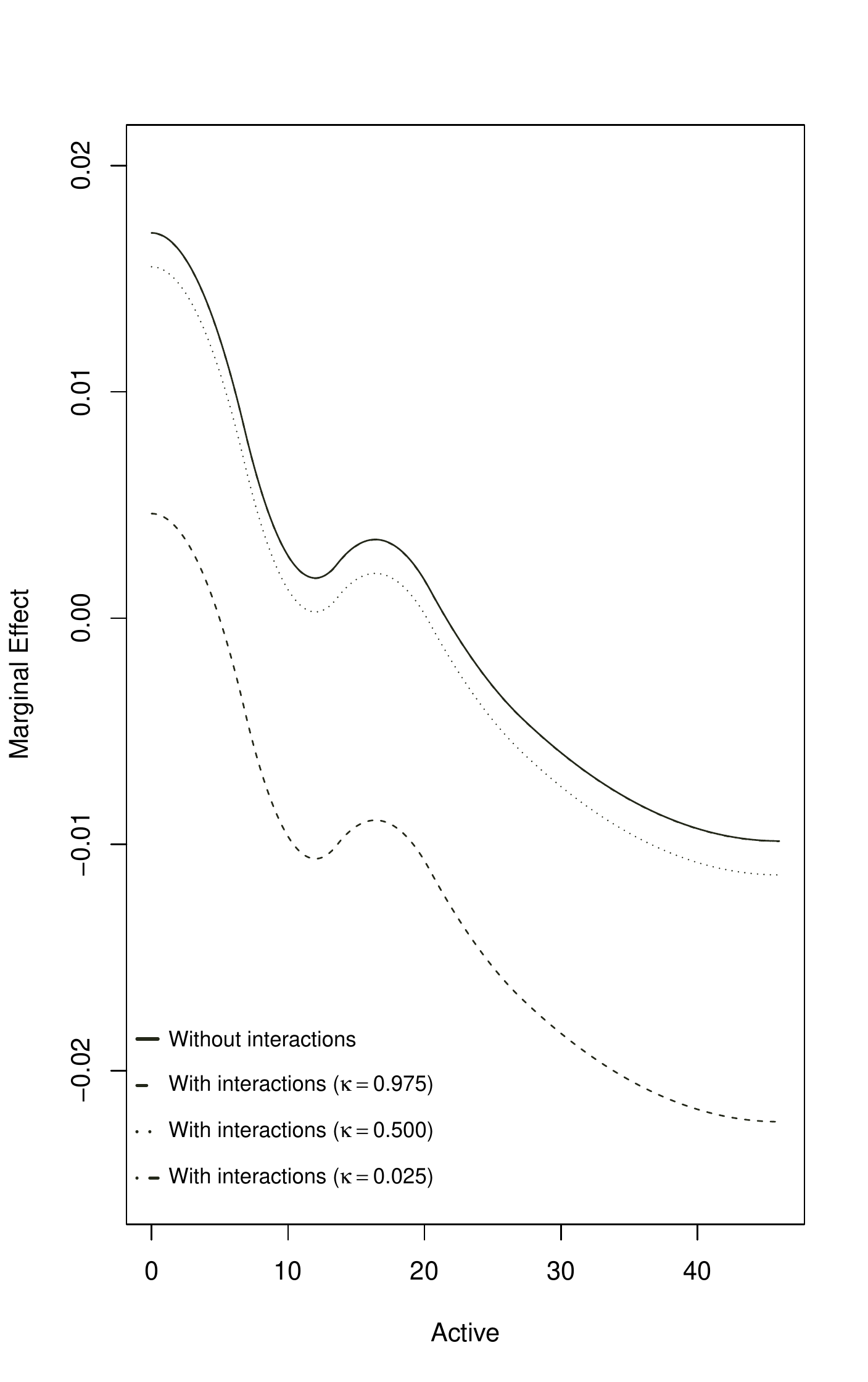}
	\vspace{-2em}
	\caption{Marginal Effect of Active (with and without interactions) for GAMA associated with the $\alpha = 0.05$ target size: credit card dataset}
	\label{fig:Marginal_Effect_Active_Translation_Interactions}
\end{figure}


As previously mentioned, the main advantage of GAM and GAM(L)A is their interpretability. 
Figure \ref{fig:Gam_functions_GAMA_lib_CreditCard} displays estimated non-linearities for GAMA associated with a target size $\alpha = 0.05$.\footnote{Table \ref{tab:CreditCard_GAM_results} in Appendix \ref{AppendixA} displays estimation results of non-parametric functions of GAMA associated with a target size $\alpha = 0.05$.} Notable among the results, a quadratic effect can be observed for the variable Active, a partial linear effect with a threshold around 3 for the variable Income, and a highly non-linear and particular effect for the variable Share. 
Figure \ref{fig:Marginal_Effect_Active_Translation_Interactions} also displays the marginal effect of the variable Active obtained for GAMA with a target size $\alpha = 0.05$. The solid curve represents the marginal effect of Active when ignoring the retained interaction variable involving Active. However, as GAMA selected one variable interacting with Active, i.e., Share, three additional curves are included in the figure. These ones represent the marginal effect of Active when Share is set to its 2.5, 50.0 and 97.5\% quantiles.\footnote{The 2.5\% quantile of variable Share is close to 0, reason why the corresponding curve displaying the marginal effects is mixed up with the solid curve (i.e., when $c_j$ is set to 0).} The graph shows that the marginal effect is non-linear and decrease with the level of Active, and that the interaction variable simply implies a change in the level of marginal effects.



\section{Conclusion}

In the wake of the growing use of machine learning (ML) algorithms in economics, interpretability has returned to the heart of the literature. Despite their high predictive performance, some ML algorithms like random forest and XGBoost are black boxes, which leads to uninterpretable models. The opacity of these algorithms has raised concerns from practitioners and regulators \Citep{Bracke2019,ACPR2020,EBA2020,EC2020} and limits their use in industry. Two approaches are currently investigated in the literature to redirect the focus to the interpretability of models. On the one hand, a first strand of the literature proposes model-agnostic approaches to improve the interpretability of black-box models. On the other hand, a second strand instead designs inherently interpretable models. While the first group of contributions has received considerably more attention in the literature, these approaches can potentially mislead users in high-stakes decisions \citep{Rudin2019}.

Against this background, we propose in this paper to rely on partial linear models that are inherently interpretable. 
Specifically, this article introduces GAM-lasso (GAMLA) and GAM-autometrics (GAMA) models, which combine parametric and non-parametric functions to accurately capture linearities and non-linearities prevailing between dependent and explanatory variables, and a variable selection procedure to control for overfitting issues. 

We propose a two-step procedure that relies on the double residual method of \citet{Robinson1988}. Monte Carlo simulation experiments show that GAMLA and GAMA can compete with benchmark ML algorithms in terms of predictive performance. Moreover, the results highlight the importance of the double residual approach as part of GAMLA and GAMA. We also recommend using GAMA because, unlike GAMLA, it allows the researcher to control the gauge level while leading to satisfying potency values.

In the empirical applications, we illustrate the predictive performance and interpretability of GAMLA and GAMA for a regression and a classification problem. Specifically, we compare GAMLA and GAMA to several other benchmark models in the literature. 
The results show that GAMLA and GAMA deliver significantly better forecasts than parametric models augmented by quadratic, cubic and interaction effects, even when lasso and autometrics are used to select the relevant models in order to account for over-fitting issues. Moreover, the results also suggest that the performance of our new models is not significantly different from that of random forest and XGBoost. We also illustrate the interpretability of GAMLA and GAMA and show that the variable selection leads to parsimonious models while graphical representations of smooth functions and marginal effects allow a simple understanding of the relationships identified by the models. 

Finally, we show in this paper that it is possible to design inherently interpretable models capable of competing with sophisticated ML algorithms like random forest and XGBoost in terms of predictive performance, and, as \Citet{Rudin2019}, we advocate for more work in this direction.


\bibliographystyle{apalike}
\bibliography{./input/Biblio_SH}

\begin{thebibliography}{}

\bibitem[ACPR, 2020]{ACPR2020}
ACPR (2020).
\newblock Governance of artificial intelligence in finance.
\newblock Discussion papers publication, November, 2020.

\bibitem[Angrist and Pischke, 2008]{Angrist2008}
Angrist, J.~D. and Pischke, J.-S. (2008).
\newblock {\em Mostly harmless econometrics}.
\newblock Princeton university press.

\bibitem[Apley and Zhu, 2020]{Apley2020}
Apley, D.~W. and Zhu, J. (2020).
\newblock Visualizing the effects of predictor variables in black box
  supervised learning models.
\newblock {\em Journal of the Royal Statistical Society: Series B (Statistical
  Methodology)}, 82(4):1059--1086.

\bibitem[Baesens et~al., 2003]{Baesens2003}
Baesens, B., Gestel, T.~V., Viaene, S., Stepanova, M., Suykens, J., and
  Vanthienen, J. (2003).
\newblock Benchmarking state-of-the-art classification algorithms for credit
  scoring.
\newblock {\em Journal of the Operational Research Society}, 54:627--635.

\bibitem[Barocas et~al., 2019]{Barocas2018}
Barocas, S., Hardt, M., and Narayanan, A. (2019).
\newblock Fairness and machine learning. fairmlbook.org, 2019.

\bibitem[Belsley et~al., 1980]{Belsley1980}
Belsley, D.~A., Kuh, E., and Welsch, R.~E. (1980).
\newblock {\em Regression diagnostics. Identifying influential data and sources
  of collinearity}.
\newblock John Wiley.

\bibitem[Boucher et~al., 2020]{Boucher2020}
Boucher, V., Bramoull{\'e}, Y., et~al. (2020).
\newblock {\em Binary Outcomes and Linear Interactions}.
\newblock Aix-Marseille School of Economics.

\bibitem[Bracke et~al., 2019]{Bracke2019}
Bracke, P., Datta, A., Jung, C., and Sen, S. (2019).
\newblock Machine learning explainability in finance: an application to default
  risk analysis.
\newblock Bank of England, Staff Working Paper No.816.

\bibitem[Breiman, 2001]{Breiman2001}
Breiman, L. (2001).
\newblock Random forests.
\newblock {\em Machine learning}, 45(1):5--32.

\bibitem[Breiman et~al., 1984]{Breiman1984}
Breiman, L., Friedman, J.~H., Olshen, R.~A., and Stone, C.~J. (1984).
\newblock {\em Classification and regression trees}.
\newblock Routledge.

\bibitem[Candelon et~al., 2012]{Candelon2012}
Candelon, B., Dumitrescu, E.-I., and Hurlin, C. (2012).
\newblock How to evaluate an early-warning system: Toward a unified statistical
  framework for assessing financial crises forecasting methods.
\newblock {\em IMF Economic Review}, 60(1):75--113.

\bibitem[Castle et~al., 2013]{Castle2013}
Castle, J.~L., Clements, M.~P., and Hendry, D.~F. (2013).
\newblock Forecasting by factors, by variables, by both or neither?
\newblock {\em Journal of Econometrics}, 177(2):305--319.

\bibitem[Castle et~al., 2011]{Castle2011}
Castle, J.~L., Doornik, J.~A., and Hendry, D.~F. (2011).
\newblock Evaluating automatic model selection.
\newblock {\em Journal of Time Series Econometrics}, 3(1).

\bibitem[Castle et~al., 2021]{Castle2021}
Castle, J.~L., Doornik, J.~A., and Hendry, D.~F. (2021).
\newblock Robust discovery of regression models.
\newblock {\em Econometrics and Statistics}.

\bibitem[Castle and Hendry, 2010]{Castle2010}
Castle, J.~L. and Hendry, D.~F. (2010).
\newblock A low-dimension portmanteau test for non-linearity.
\newblock {\em Journal of Econometrics}, 158(2):231--245.

\bibitem[Charpentier et~al., 2018]{Charpentier2018}
Charpentier, A., Flachaire, E., and Ly, A. (2018).
\newblock Econometrics and machine learning.
\newblock {\em Economie et Statistique}, 505(1):147--169.

\bibitem[Chen et~al., 2015]{Chen2015}
Chen, T., He, T., Benesty, M., Khotilovich, V., Tang, Y., Cho, H., et~al.
  (2015).
\newblock Xgboost: extreme gradient boosting.
\newblock R package version 0.4-2.

\bibitem[Chouldechova and Hastie, 2015]{Chouldechova2015}
Chouldechova, A. and Hastie, T. (2015).
\newblock Generalized additive model selection.
\newblock {\em preprint arXiv:1506.03850}.

\bibitem[Desai et~al., 1996]{Desai1996}
Desai, V.~S., Crook, J.~N., and Overstreet~Jr, G.~A. (1996).
\newblock A comparison of neural networks and linear scoring models in the
  credit union environment.
\newblock {\em European Journal of Operational Research}, 95(1):24--37.

\bibitem[Doornik, 2009]{doornik2009autometrics}
Doornik, J.~A. (2009).
\newblock Autometrics. in {J}.{L} {C}astle and {N}. {S}hepards ({E}ds.).
\newblock {\em The Methodology and Practice of Econometrics: Festschrift in
  Honour of David F. Hendry.}
\newblock 88--121.

\bibitem[Doornik and Hansen, 2008]{Doornik2008}
Doornik, J.~A. and Hansen, H. (2008).
\newblock An omnibus test for univariate and multivariate normality.
\newblock {\em Oxford bulletin of economics and statistics}, 70:927--939.

\bibitem[Doshi-Velez and Kim, 2017]{Doshi2017}
Doshi-Velez, F. and Kim, B. (2017).
\newblock Towards a rigorous science of interpretable machine learning.
\newblock {\em preprint arXiv:1702.08608}.

\bibitem[Dumitrescu et~al., 2022]{Dumitrescu2022}
Dumitrescu, E., Hue, S., Hurlin, C., and Tokpavi, S. (2022).
\newblock Machine learning for credit scoring: Improving logistic regression
  with non-linear decision-tree effects.
\newblock {\em European Journal of Operational Research}, 297(3):1178--1192.

\bibitem[EBA, 2020]{EBA2020}
EBA (2020).
\newblock Report on big data and advanced analytics.
\newblock European Banking Authority, January, 2020.

\bibitem[EC, 2020]{EC2020}
EC (2020).
\newblock White paper on artificial intelligence: A european approach to
  excellence and trust.
\newblock European Commission, February, 2020.

\bibitem[Engle, 1982]{Engle1982}
Engle, R.~F. (1982).
\newblock Autoregressive conditional heteroscedasticity with estimates of the
  variance of united kingdom inflation.
\newblock {\em Econometrica}, 50(4).
\newblock 987--1007.

\bibitem[Epprecht et~al., 2021]{Epprecht2021}
Epprecht, C., Guegan, D., Veiga, {\'A}., and Correa~da Rosa, J. (2021).
\newblock Variable selection and forecasting via automated methods for linear
  models: Lasso/adalasso and autometrics.
\newblock {\em Communications in Statistics-Simulation and Computation},
  50(1):103--122.

\bibitem[Fan and Li, 2001]{Fan2001}
Fan, J. and Li, R. (2001).
\newblock Variable selection via nonconcave penalized likelihood and its oracle
  properties.
\newblock {\em Journal of the American statistical Association},
  96(456):1348--1360.

\bibitem[Finlay, 2011]{Finlay2011}
Finlay, S. (2011).
\newblock Multiple classifier architectures and their application to credit
  risk assessment.
\newblock {\em European Journal of Operational Research}, 210(2):368--378.

\bibitem[Friedman, 2001]{Friedman2001}
Friedman, J.~H. (2001).
\newblock Greedy function approximation: A gradient boosting machine.
\newblock {\em The Annals of Statistics}, 29(5):1189--1232.

\bibitem[Frisch and Waugh, 1933]{Frisch1933}
Frisch, R. and Waugh, F.~V. (1933).
\newblock Partial time regressions as compared with individual trends.
\newblock {\em Econometrica}, 1(4).
\newblock 387--401.

\bibitem[Ghosal and Kormaksson, 2019]{plsmselect}
Ghosal, I. and Kormaksson, M. (2019).
\newblock The plsmselect package.
\newblock
  https://cran.r-project.org/web/packages/plsmselect/vignettes/plsmselect.html.

\bibitem[Godfrey, 1978]{Godfrey1978}
Godfrey, L.~G. (1978).
\newblock Testing for higher order serial correlation in regression equations
  when the regressors include lagged dependent variables.
\newblock {\em Econometrica}, 46(6).
\newblock 1303--1310.

\bibitem[Greene, 2003]{greene2003econometric}
Greene, W.~H. (2003).
\newblock {\em Econometric analysis}.
\newblock Pearson Education India.

\bibitem[Grennepois et~al., 2018]{Grennepois2018}
Grennepois, N., Alvirescu, M., and Bombail, M. (2018).
\newblock Using random forest for credit risk models.
\newblock Deloitte Risk Advisory, September.

\bibitem[Gunnarsson et~al., 2021]{Gunnarsson2021}
Gunnarsson, B.~R., Vanden~Broucke, S., Baesens, B., {\'O}skarsd{\'o}ttir, M.,
  and Lemahieu, W. (2021).
\newblock Deep learning for credit scoring: Do or don't?
\newblock {\em European Journal of Operational Research}, 295(1):292--305.

\bibitem[Hansen et~al., 2011]{Hansen2011}
Hansen, P.~R., Lunde, A., and Nason, J.~M. (2011).
\newblock The model confidence set.
\newblock {\em Econometrica}, 79(2):453--497.

\bibitem[Harrison~Jr and Rubinfeld, 1978]{harrison1978hedonic}
Harrison~Jr, D. and Rubinfeld, D.~L. (1978).
\newblock Hedonic housing prices and the demand for clean air.
\newblock {\em Journal of environmental economics and management}, 5:81--102.

\bibitem[Hastie and Tibshirani, 2017]{Hastie2017}
Hastie, T.~J. and Tibshirani, R.~J. (2017).
\newblock {\em Generalized additive models}.
\newblock Routledge.

\bibitem[Hendry, 2000]{Hendry2000}
Hendry, D.~F. (2000).
\newblock {\em Econometrics: alchemy or science?: essays in econometric
  methodology}.
\newblock Oxford University Press.

\bibitem[Hendry and Doornik, 2014]{Hendry2014}
Hendry, D.~F. and Doornik, J.~A. (2014).
\newblock {\em Empirical model discovery and theory evaluation: automatic
  selection methods in econometrics}.
\newblock MIT Press.

\bibitem[Hendry and Johansen, 2011]{Hendry2011}
Hendry, D.~F. and Johansen, S. (2011).
\newblock The properties of model selection when retaining theory variables.
\newblock {\em University of Copenhagen Discussion Paper}.

\bibitem[Henley and Hand, 1996]{Henley1996}
Henley, W. and Hand, D. (1996).
\newblock A k-nearest-neighbour classifier for assessing consumer credit risk.
\newblock {\em The Statistician}, 45(1):77--95.

\bibitem[Hurlin and P{\'e}rignon, 2019]{Hurlin2019}
Hurlin, C. and P{\'e}rignon, C. (2019).
\newblock Machine learning et nouvelles sources de donn{\'e}es pour le scoring
  de cr{\'e}dit.
\newblock {\em Revue d'{\'e}conomie financi{\`e}re}, (3):21--50.

\bibitem[Hurlin et~al., 2021]{Hurlin2021}
Hurlin, C., P{\'e}rignon, C., and Saurin, S. (2021).
\newblock The fairness of credit scoring models.
\newblock {\em Available at SSRN 3785882}.

\bibitem[Johansen and Nielsen, 2016]{Johansen2016}
Johansen, S. and Nielsen, B. (2016).
\newblock Asymptotic theory of outlier detection algorithms for linear time
  series regression models.
\newblock {\em Scandinavian Journal of Statistics}, 43(2):321--348.

\bibitem[Kozodoi et~al., 2022]{Kozodoi2021}
Kozodoi, N., Jacob, J., and Lessmann, S. (2022).
\newblock Fairness in credit scoring: Assessment, implementation and profit
  implications.
\newblock {\em European Journal of Operational Research}, 297(3):1083--1094.

\bibitem[Lessmann et~al., 2015]{Lessmann2015}
Lessmann, S., Baesens, B., Seow, H.-V., and Thomas, L.~C. (2015).
\newblock Benchmarking state-of-the-art classification algorithms for credit
  scoring: An update of research.
\newblock {\em European Journal of Operational Research}, 247:124--136.

\bibitem[Li and Racine, 2007]{Li2007}
Li, Q. and Racine, J.~S. (2007).
\newblock {\em Nonparametric econometrics: theory and practice}.
\newblock Princeton University Press.

\bibitem[Lovell, 1963]{Lovell1963}
Lovell, M.~C. (1963).
\newblock Seasonal adjustment of economic time series and multiple regression
  analysis.
\newblock {\em Journal of the American Statistical Association},
  58(304):993--1010.

\bibitem[Lundberg and Lee, 2017]{Lundberg2017}
Lundberg, S.~M. and Lee, S.-I. (2017).
\newblock A unified approach to interpreting model predictions.
\newblock In {\em Proceedings of the 31st international conference on neural
  information processing systems}.
\newblock 4768--4777.

\bibitem[Makowski, 1985]{Makowski1985}
Makowski, P. (1985).
\newblock Credit scoring branches out.
\newblock {\em Credit World}, 75(1):30--37.

\bibitem[Michelucci and Venturini, 2021]{Michelucci2021}
Michelucci, U. and Venturini, F. (2021).
\newblock Estimating neural network's performance with bootstrap: A tutorial.
\newblock {\em Machine Learning and Knowledge Extraction}, 3(2):357--373.

\bibitem[Molnar, 2019]{Molnar2019}
Molnar, C. (2019).
\newblock Interpretable machine learning: A guide for making black box models
  explainable. published online.

\bibitem[Molnar et~al., 2020]{Molnar2020}
Molnar, C., Casalicchio, G., and Bischl, B. (2020).
\newblock Interpretable machine learning--a brief history, state-of-the-art and
  challenges.
\newblock In {\em Joint European Conference on Machine Learning and Knowledge
  Discovery in Databases}. Springer.
\newblock 417--431.

\bibitem[Paleologo et~al., 2010]{Paleologo2010}
Paleologo, G., Elisseeff, A., and Antonini, G. (2010).
\newblock Subagging for credit scoring models.
\newblock {\em European Journal of Operational Research}, 201(2):490--499.

\bibitem[Ramsey, 1969]{Ramsey1969}
Ramsey, J.~B. (1969).
\newblock Tests for specification errors in classical linear least-squares
  regression analysis.
\newblock {\em Journal of the Royal Statistical Society: Series B
  (Methodological)}, 31(2):350--371.

\bibitem[Ribeiro et~al., 2016]{Ribeiro2016}
Ribeiro, M.~T., Singh, S., and Guestrin, C. (2016).
\newblock " {W}hy should {I} trust you?" explaining the predictions of any
  classifier.
\newblock In {\em Proceedings of the 22nd ACM SIGKDD international conference
  on knowledge discovery and data mining}.
\newblock 1135--1144.

\bibitem[Robinson, 1988]{Robinson1988}
Robinson, P.~M. (1988).
\newblock Root-n-consistent semiparametric regression.
\newblock {\em Econometrica:}, 56(4).
\newblock 931--954.

\bibitem[Rudin, 2019]{Rudin2019}
Rudin, C. (2019).
\newblock Stop explaining black box machine learning models for high stakes
  decisions and use interpretable models instead.
\newblock {\em Nature Machine Intelligence}, 1(5):206--215.

\bibitem[Rudin et~al., 2021]{Rudin2021}
Rudin, C., Chen, C., Chen, Z., Huang, H., Semenova, L., and Zhong, C. (2021).
\newblock Interpretable machine learning: Fundamental principles and 10 grand
  challenges.
\newblock {\em arXiv preprint arXiv:2103.11251}.

\bibitem[Rudin and Radin, 2019]{Rudin2019b}
Rudin, C. and Radin, J. (2019).
\newblock Why are we using black box models in {AI} when we don't need to? {A}
  lesson from an explainable {AI} competition.
\newblock {\em Harvard Data Science Review}, 1(2):1--9.

\bibitem[Tibshirani, 1996]{Tibshirani1996}
Tibshirani, R. (1996).
\newblock Regression shrinkage and selection via the lasso.
\newblock {\em Journal of the Royal Statistical Society: Series B
  (Methodological)}, 58(1):267--288.

\bibitem[Varian, 2014]{Varian2014}
Varian, H.~R. (2014).
\newblock Big data: New tricks for econometrics.
\newblock {\em Journal of Economic Perspectives}, 28(2):3--28.

\bibitem[White, 1980]{White1980}
White, H. (1980).
\newblock A heteroskedasticity-consistent covariance matrix estimator and a
  direct test for heteroskedasticity.
\newblock {\em Econometrica}, 48(4).
\newblock 817--838.

\bibitem[Wooldridge, 2015]{Wooldridge2015}
Wooldridge, J.~M. (2015).
\newblock {\em Introductory econometrics: A modern approach}.
\newblock Cengage learning.

\bibitem[Zou, 2006]{Zou2006}
Zou, H. (2006).
\newblock The adaptive lasso and its oracle properties.
\newblock {\em Journal of the American Statistical Association},
  101(476):1418--1429.

\end{thebibliography}

\newpage 

\appendix

\section{Additional Figures and Tables}\label{AppendixA}

\renewcommand{\thetable}{T\arabic{table}}
\renewcommand{\thefigure}{F\arabic{figure}}


\begin{table}[!htbp] \centering 
	\caption{Comparison of potency, gauge and MSE under non-linear effects and uncorrelated covariates} 
	\renewcommand{\arraystretch}{1.6}
	\begin{threeparttable}
		\begin{tabular}{lccccc} 
			\hline
			Model & Conditional mean & Tuning parameter  & Potency & Gauge & MSE \\
			\hline
			OLS & $(X^1,X^2,X^3,I)\beta$ & & & & $1.154$ \\ 
			LASSO & $(X^1,X^2,X^3,I)\beta$ & $\hat{\lambda}^{min}$ & $1$ & $0.369$ & $1.104$ \\ 
			& $(X^1,X^2,X^3,I)\beta$ & $\hat{\lambda}^{1se}$ & $0.998$ & $0.040$ & $1.145$ \\ 
			A-LASSO & $(X^1,X^2,X^3,I)\beta$ & $\hat{\lambda}^{min}$ & $0.998$ & $0.039$ & $1.102$ \\ 
			& $(X^1,X^2,X^3,I)\beta$ & $\hat{\lambda}^{1se}$ & $0.983$ & $0.005$ & $1.130$ \\ 
			AM & $(X^1,X^2,X^3,I)\beta$ & $\alpha = 0.05$ & $1$ & $0.050$ & $1.103$ \\ 
			& $(X^1,X^2,X^3,I)\beta$ & $\alpha = 0.01$ & $1$ & $0.011$ & $1.095$ \\ 
			GAMLA* & $I\gamma + \sum_{j=1}^{p}g_j\left(X_j\right)$ & $\hat{\lambda}^{min}$ & $1$ & $0.252$ & $1.091$ \\ 
			& $I\gamma + \sum_{j=1}^{p}g_j\left(X_j\right)$ & $\hat{\lambda}^{1se}$ & $0.990$ & $0.007$ & $1.062$ \\ 
			GAMLA* (A-LASSO) & $I\gamma + \sum_{j=1}^{p}g_j\left(X_j\right)$ & $\hat{\lambda}^{min}$ & $0.990$ & $0.007$ & $1.062$ \\ 
			& $I\gamma + \sum_{j=1}^{p}g_j\left(X_j\right)$ & $\hat{\lambda}^{1se}$ & $0.911$ & $0.000$ & $1.075$ \\ 
			GAMA* & $I\gamma + \sum_{j=1}^{p}g_j\left(X_j\right)$ & $\alpha = 0.05$ & $1$ & $0.053$ & $1.072$ \\ 
			& $I\gamma + \sum_{j=1}^{p}g_j\left(X_j\right)$ & $\alpha = 0.01$ & $1$ & $0.010$ & $1.061$ \\ 
			GAMLA & $I\gamma + \sum_{j=1}^{p}g_j\left(X_j\right)$ & $\hat{\lambda}^{min}$ & $1$ & $0.273$ & $1.093$ \\ 
			& $I\gamma + \sum_{j=1}^{p}g_j\left(X_j\right)$ & $\hat{\lambda}^{1se}$ & $0.992$ & $0.008$ & $1.062$ \\ 
			GAMLA (A-LASSO) & $I\gamma + \sum_{j=1}^{p}g_j\left(X_j\right)$ & $\hat{\lambda}^{min}$ & $0.991$ & $0.007$ & $1.062$ \\ 
			& $I\gamma + \sum_{j=1}^{p}g_j\left(X_j\right)$ & $\hat{\lambda}^{1se}$ & $0.923$ & $0.000$ & $1.073$ \\ 
			GAMA & $I\gamma + \sum_{j=1}^{p}g_j\left(X_j\right)$ & $\alpha = 0.05$ & $1$ & $0.057$ & $1.073$ \\ 
			& $I\gamma + \sum_{j=1}^{p}g_j\left(X_j\right)$ & $\alpha = 0.01$ & $1$ & $0.013$ & $1.062$ \\ 
			GAMSEL & $f_{Gs}\left(X\right)$ & $\hat{\lambda}^{min}$ & $1$ & $0.473$ & $1.107$ \\ 
			& $f_{Gs}\left(X\right)$ & $\hat{\lambda}^{1se}$ & $1$ & $0.063$ & $1.140$ \\ 
			Random Forest & $f_{Rf}\left(X\right)$ & & & & $1.200$ \\ 
			XGBoost & $f_{Gb}\left(X\right)$ & & & & $1.287$ \\ 
			\hline
		\end{tabular} 
		\begin{tablenotes}[para,flushleft]
			\small
			\noindent Note: See Table \ref{Simu_Setup1}.
		\end{tablenotes}
	\end{threeparttable}
	\label{Simu_Setup2}
\end{table} 


\begin{table}[!htbp] \centering 
	\caption{Comparison of potency, gauge and MSE under linear effects and correlated covariates} 
	\renewcommand{\arraystretch}{1.6}
	\begin{threeparttable}
		\begin{tabular}{lccccc} 
			\hline
			Model & Conditional mean & Tuning parameter  & Potency & Gauge & MSE \\
			\hline
			OLS & $(X^1,X^2,X^3,I)\beta$ & & & & $1.093$ \\ 
			LASSO & $(X^1,X^2,X^3,I)\beta$ & $\hat{\lambda}^{min}$ & $1$ & $0.641$ & $1.061$ \\ 
			& $(X^1,X^2,X^3,I)\beta$ & $\hat{\lambda}^{1se}$ & $0.991$ & $0.462$ & $1.106$ \\ 
			A-LASSO & $(X^1,X^2,X^3,I)\beta$ & $\hat{\lambda}^{min}$ & $0.988$ & $0.198$ & $1.035$ \\ 
			& $(X^1,X^2,X^3,I)\beta$ & $\hat{\lambda}^{1se}$ & $0.935$ & $0.040$ & $1.066$ \\ 
			AM & $(X^1,X^2,X^3,I)\beta$ & $\alpha = 0.05$ & $1$ & $0.047$ & $1.028$ \\ 
			& $(X^1,X^2,X^3,I)\beta$ & $\alpha = 0.01$ & $1$ & $0.010$ & $1.017$ \\ 
			GAMLA* & $I\gamma + \sum_{j=1}^{p}g_j\left(X_j\right)$ & $\hat{\lambda}^{min}$ & $1$ & $0.753$ & $1.081$ \\ 
			& $I\gamma + \sum_{j=1}^{p}g_j\left(X_j\right)$ & $\hat{\lambda}^{1se}$ & $0.950$ & $0.371$ & $1.076$ \\ 
			GAMLA* (A-LASSO) & $I\gamma + \sum_{j=1}^{p}g_j\left(X_j\right)$ & $\hat{\lambda}^{min}$ & $0.974$ & $0.344$ & $1.071$ \\ 
			& $I\gamma + \sum_{j=1}^{p}g_j\left(X_j\right)$ & $\hat{\lambda}^{1se}$ & $0.864$ & $0.159$ & $1.083$ \\ 
			GAMA* & $I\gamma + \sum_{j=1}^{p}g_j\left(X_j\right)$ & $\alpha = 0.05$ & $0.996$ & $0.212$ & $1.059$ \\ 
			& $I\gamma + \sum_{j=1}^{p}g_j\left(X_j\right)$ & $\alpha = 0.01$ & $0.969$ & $0.153$ & $1.060$ \\ 
			GAMLA & $I\gamma + \sum_{j=1}^{p}g_j\left(X_j\right)$ & $\hat{\lambda}^{min}$ & $1$ & $0.263$ & $1.070$ \\ 
			& $I\gamma + \sum_{j=1}^{p}g_j\left(X_j\right)$ & $\hat{\lambda}^{1se}$ & $0.995$ & $0.008$ & $1.041$ \\ 
			GAMLA (A-LASSO) & $I\gamma + \sum_{j=1}^{p}g_j\left(X_j\right)$ & $\hat{\lambda}^{min}$ & $0.995$ & $0.008$ & $1.041$ \\ 
			& $I\gamma + \sum_{j=1}^{p}g_j\left(X_j\right)$ & $\hat{\lambda}^{1se}$ & $0.928$ & $0.000$ & $1.051$ \\ 
			GAMA & $I\gamma + \sum_{j=1}^{p}g_j\left(X_j\right)$ & $\alpha = 0.05$ & $1$ & $0.054$ & $1.051$ \\ 
			& $I\gamma + \sum_{j=1}^{p}g_j\left(X_j\right)$ & $\alpha = 0.01$ & $1$ & $0.012$ & $1.042$ \\ 
			GAMSEL & $f_{Gs}\left(X\right)$ & $\hat{\lambda}^{min}$ & $1$ & $0.711$ & $1.076$ \\ 
			& $f_{Gs}\left(X\right)$ & $\hat{\lambda}^{1se}$ & $0.986$ & $0.517$ & $1.112$ \\ 
			Random Forest & $f_{Rf}\left(X\right)$ & & & & $1.419$ \\ 
			XGBoost & $f_{Gb}\left(X\right)$ & & & & $1.217$ \\ 
			\hline
		\end{tabular} 
		\begin{tablenotes}[para,flushleft]
			\small
			\noindent Note: See Table \ref{Simu_Setup1}.
			\end{tablenotes}
	\end{threeparttable}
	\label{Simu_Setup3}
\end{table}


\begin{table}[!htbp] \centering 
	\caption{Comparison of potency, gauge and MSE under linear effects and uncorrelated covariates} 
	\renewcommand{\arraystretch}{1.6}
	\begin{threeparttable}
		\begin{tabular}{lccccc} 
			\hline
			Model & Conditional mean & Tuning parameter  & Potency & Gauge & MSE \\
			\hline
			OLS & $(X^1,X^2,X^3,I)\beta$ & & & & $1.089$ \\ 
			LASSO & $(X^1,X^2,X^3,I)\beta$ & $\hat{\lambda}^{min}$ & $1$ & $0.298$ & $1.034$ \\ 
			& $(X^1,X^2,X^3,I)\beta$ & $\hat{\lambda}^{1se}$ & $1$ & $0.032$ & $1.073$ \\ 
			A-LASSO & $(X^1,X^2,X^3,I)\beta$ & $\hat{\lambda}^{min}$ & $1$ & $0.029$ & $1.020$ \\ 
			& $(X^1,X^2,X^3,I)\beta$ & $\hat{\lambda}^{1se}$ & $0.991$ & $0.001$ & $1.053$ \\ 
			AM & $(X^1,X^2,X^3,I)\beta$ & $\alpha = 0.05$ & $1$ & $0.049$ & $1.030$ \\ 
			& $(X^1,X^2,X^3,I)\beta$ & $\alpha = 0.01$ & $1$ & $0.011$ & $1.017$ \\ 
			GAMLA* & $I\gamma + \sum_{j=1}^{p}g_j\left(X_j\right)$ & $\hat{\lambda}^{min}$ & $1$ & $0.239$ & $1.061$ \\ 
			& $I\gamma + \sum_{j=1}^{p}g_j\left(X_j\right)$ & $\hat{\lambda}^{1se}$ & $0.994$ & $0.006$ & $1.034$ \\ 
			GAMLA* (A-LASSO) & $I\gamma + \sum_{j=1}^{p}g_j\left(X_j\right)$ & $\hat{\lambda}^{min}$ & $0.994$ & $0.007$ & $1.034$ \\ 
			& $I\gamma + \sum_{j=1}^{p}g_j\left(X_j\right)$ & $\hat{\lambda}^{1se}$ & $0.924$ & $0.000$ & $1.045$ \\ 
			GAMA* & $I\gamma + \sum_{j=1}^{p}g_j\left(X_j\right)$ & $\alpha = 0.05$ & $1$ & $0.050$ & $1.043$ \\ 
			& $I\gamma + \sum_{j=1}^{p}g_j\left(X_j\right)$ & $\alpha = 0.01$ & $1$ & $0.011$ & $1.034$ \\ 
			GAMLA & $I\gamma + \sum_{j=1}^{p}g_j\left(X_j\right)$ & $\hat{\lambda}^{min}$ & $1$ & $0.258$ & $1.062$ \\ 
			& $I\gamma + \sum_{j=1}^{p}g_j\left(X_j\right)$ & $\hat{\lambda}^{1se}$ & $0.992$ & $0.008$ & $1.035$ \\
			GAMLA (A-LASSO) & $I\gamma + \sum_{j=1}^{p}g_j\left(X_j\right)$ & $\hat{\lambda}^{min}$ & $0.993$ & $0.007$ & $1.034$ \\ 
			& $I\gamma + \sum_{j=1}^{p}g_j\left(X_j\right)$ & $\hat{\lambda}^{1se}$ & $0.926$ & $0.000$ & $1.045$ \\  
			GAMA & $I\gamma + \sum_{j=1}^{p}g_j\left(X_j\right)$ & $\alpha = 0.05$ & $1$ & $0.053$ & $1.044$ \\ 
			& $I\gamma + \sum_{j=1}^{p}g_j\left(X_j\right)$ & $\alpha = 0.01$ & $1$ & $0.012$ & $1.035$ \\ 
			GAMSEL & $f_{Gs}\left(X\right)$ & $\hat{\lambda}^{min}$ & $1$ & $0.361$ & $1.030$ \\ 
			& $f_{Gs}\left(X\right)$ & $\hat{\lambda}^{1se}$ & $1$ & $0.045$ & $1.060$ \\ 
			Random Forest & $f_{Rf}\left(X\right)$ & & & & $1.174$ \\ 
			XGBoost & $f_{Gb}\left(X\right)$ & & & & $1.212$ \\ 
			\hline
		\end{tabular} 
		\begin{tablenotes}[para,flushleft]
			\small
			\noindent Note: See Table \ref{Simu_Setup1}.
	\end{tablenotes}
	\end{threeparttable}
	\label{Simu_Setup4}
\end{table}

 \begin{table}[!htbp] 
 	\centering 
	\caption{Description of the variables in the Boston dataset}
	\begin{threeparttable}
	\begin{tabular}{l|l}
		\hline
		Variable & Description\\
		\hline
		Medv & Median value of owner-occupied homes in \$1000s \\
		Crim & Per capita crime rate by town \\
		Zn & Proportion of residential land zoned for lots over 25,000 square feet\\
		Indus & Proportion of non-retail business acres per town\\
		Chas & Charles River dummy variable (= 1 if tract bounds river, 0 otherwise) \\
		Nox & Nitric oxides concentration (parts per 10 million)\\
		Rm & Average number of rooms per dwelling\\
		Age & Proportion of owner-occupied units built prior to 1940\\
		Dis & Weighted distances to five Boston employment centres\\
		Rad & Index of accessibility to radial highways \\
		Tax & Full-value property-tax rate per \$10,000\\
		Ptratio & Pupil-teacher ratio by town\\
		Black & $1000(B_k - 0.63)^2$, where $B_k$ is the
                        proportion of black persons by town\\
		Lstat & \% Lower status of the population\\
		\hline
	\end{tabular}
	\begin{tablenotes}[para,flushleft]
		\small
		\item Note: See \citet{harrison1978hedonic} for more details on the dataset.
	\end{tablenotes}
	\end{threeparttable}
\label{tab:dicboston}
\end{table}

 \begin{table}[!htbp]
 	\centering 
	\caption{Description of the variables in the credit card dataset}
	\begin{threeparttable}
	\begin{tabular}{l|l}
		\hline 
		Variable & Description \\
		\hline 
		Card & Dummy variable: 1 if application for credit card accepted, 0 if not \\
		Reports & Number of major derogatory reports \\
		Age & Age in years plus twelfths of a year \\
		Income & Yearly income (in USD 10,000) \\
		Share & Ratio of monthly credit card expenditure to yearly income \\
		Expenditure & Average monthly credit card expenditure \\
		Owner &  Dummy variable: 1 if owns their home, 0 if rent \\
		Selfemp  &  Dummy variable: 1 if self employed, 0 if not \\
		Dependents & Number of dependents \\
		Months & Months living at current address \\
		Majorcards & Number of major credit cards held (0 or 1) \\
		Active & Number of active credit accounts \\
		\hline 
	\end{tabular}
	\begin{tablenotes}[para,flushleft]
		\small
		\item Note: See \cite{greene2003econometric}	 for more details on the dataset.
	\end{tablenotes}
	\end{threeparttable}
	\label{tab:cred}
\end{table}

 \begin{table}[!htbp]
	\centering 
	\caption{Estimation results of non-parametric functions of GAMA associated with a target size $\alpha = 0.05$: Boston housing dataset}	
	\begin{threeparttable}
    \begin{tabular}{l|cccr}
    	\hline
    	Variable & Edf & Df & F-stat & P-value \\
		\hline
		Crim & $4.488$ & $5$ & $6.870$ & $<0.001$ \\
		Zn & $0.000$ & $5$ & $0.000$ & $0.681$ \\
		Indus & $1.840$ & $5$ & $2.308$ & $0.001$ \\
		Nox & $0.235$ & $5$ & $0.067$ & $0.200$ \\
		Rm & $3.997$ & $5$ & $55.557$ & $<0.001$ \\
		Age & $4.164$ & $5$ & $3.271$ & $0.002$ \\
		Dis & $4.974$ & $5$ & $26.247$ & $<0.001$ \\
		Tax & $4.302$ & $5$ & $4.201$ & $<0.001$ \\
		Ptratio & $0.000$ & $5$ & $0.000$ & $0.458$ \\
		Black & $4.705$ & $5$ & $11.481$ & $<0.001$ \\
		Lstat & $4.551$ & $5$ & $15.263$ & $<0.001$ \\
		\hline
	\end{tabular}
	\begin{tablenotes}[para,flushleft]
	\small
	\item Note: This table displays estimation results of non-parametric functions of GAMA associated with a target size $\alpha = 0.05$ for the Boston housing dataset.
	\end{tablenotes}
	\end{threeparttable}
	\label{tab:Boston_GAM_results}
\end{table}

\begin{table}[htbp] 
	\centering 
	\caption{Number of variables selected, MSE and MCS: Boston housing dataset with a dummy variable for Boston and its suburbs} 
	\footnotesize
	\renewcommand{\arraystretch}{1.6}
	\begin{adjustwidth}{-1cm}{}
		\begin{threeparttable}
			\begin{tabular}{llccccr}
				\hline
				\# & Model & Conditional  & Tuning  & Number of interactions & MSE & MCS P-value \\  
				 & & mean & parameter &   & &  \\  
				\hline
				$[1]$ & GAMLA & $I\gamma + \sum_{j=1}^{p}g_j\left(X_j\right)$ & $\hat{\lambda}^{min}$ & 64 & $10.247$ & $0.936$ \\ 
				&GAMLA & $I\gamma + \sum_{j=1}^{p}g_j\left(X_j\right)$ & $\hat{\lambda}^{1se}$ & 25 & $9.700$ & $1.000$ \\ 
				$[2]$ &GAMA & $I\gamma + \sum_{j=1}^{p}g_j\left(X_j\right)$ & $\alpha = 0.05$ & 26 & $9.928$ & $0.984$ \\ 
				&GAMA & $I\gamma + \sum_{j=1}^{p}g_j\left(X_j\right)$ & $\alpha = 0.01$ & 17 & $10.169$ & $0.936$ \\  
				\hline
				$[3]$ & GAMLA (A-LASSO) & $I\gamma + \sum_{j=1}^{p}g_j\left(X_j\right)$ & $\hat{\lambda}^{min}$ & 18 & $10.134$ & $0.936$ \\ 
				&GAMLA (A-LASSO) & $I\gamma + \sum_{j=1}^{p}g_j\left(X_j\right)$ & $\hat{\lambda}^{1se}$ & 9 & $11.004$ & $0.593$ \\ 
				\hline
				$[4]$ &OLS & $X\beta$ & & & $23.984$ & $<0.001$ \\ 
				$[5]$ &OLS & $(X^1,X^2,X^3)\beta$ & & & $16.030$ & $0.006$ \\ 
				$[6]$ &OLS & $(X^1,X^2,X^3,I)\beta$ & & 91 & $23.140$ & $0.006$ \\ 
				\hline
				$[7]$ &GAM & $\sum_{j=1}^{p}g_j\left(X_j\right)$ & & & $13.106$ & $0.008$ \\ 
				\hline
				$[8]$ &LASSO & $(X^1,X^2,X^3,I)\beta$ & $\hat{\lambda}^{min}$ & 78 & $14.056$ & $0.101$ \\ 
				&LASSO & $(X^1,X^2,X^3,I)\beta$ & $\hat{\lambda}^{1se}$ & 37 & $16.234$ & $0.006$ \\ 
				$[9]$ &A-LASSO & $(X^1,X^2,X^3,I)\beta$ & $\hat{\lambda}^{min}$ & 14 & $22.242$ & $0.006$ \\ 
				&A-LASSO & $(X^1,X^2,X^3,I)\beta$ & $\hat{\lambda}^{1se}$ & 10 & $26.087$ & $0.004$ \\ 
				$[10]$ &AM & $(X^1,X^2,X^3,I)\beta$ & $\alpha = 0.05$ & 30 & $16.631$ & $0.007$ \\ 
				&AM & $(X^1,X^2,X^3,I)\beta$ & $\alpha = 0.01$ & 32 & $15.599$ & $0.006$ \\ 
				\hline
				$[11]$ &Random Forest & $f_{Rf}\left(X\right)$ & & & $10.075$ & $0.984$ \\ 
				$[12]$ &XGBoost & $f_{Gb}\left(X\right)$ & & & $9.726$ & $0.985$ \\
				\hline
				$[13]$ &PLTR & $X\beta + \mathcal{V}_1\xi + \mathcal{V}_2\zeta$ & &  & $13.062$ & $0.024$ \\
				\hline
			\end{tabular} 
			\begin{tablenotes}[para,flushleft]
				\small
				\noindent Note: The conditional mean of the competing models/estimation methods is provided in the second column (labeled `Conditional mean').  The results displayed in column MCS correspond to the mean square (forecast) error over the 10-folds while those in the last column are the p-values of the MCS test based on 10,000 bootstrap samples. 
			\end{tablenotes}
		\end{threeparttable}
	\end{adjustwidth}
	\label{Boston_results_dummy} 
\end{table} 

 \begin{table}[!htbp]
	\centering 
	\caption{Estimation results of non-parametric functions of GAMA associated with the $\alpha = 0.05$ target size: Credit card dataset}	
	\begin{threeparttable}
    \begin{tabular}{l|cccr}
		\hline
		Variable & Edf & Df & F-stat & P-value \\
		\hline
		Reports & $3.222$ & $5$ & $35.754$ & $<0.001$ \\
		Age & $2.715$ & $5$ & $1.595$ & $0.023$ \\
		Income & $3.334$ & $5$ & $9.451$ & $<0.001$ \\
		Share & $5.000$ & $5$ & $84.070$ & $<0.001$ \\
		Expenditure & $4.770$ & $5$ & $5.821$ & $<0.001$ \\
		Dependents & $1.120$ & $5$ & $2.425$ & $<0.001$ \\
		Months & $2.070$ & $5$ & $0.990$ & $0.067$ \\
		Active & $3.680$ & $5$ & $10.915$ & $<0.001$ \\
		\hline
	\end{tabular}
	\begin{tablenotes}[para,flushleft]
	\small
	\item Note: This table displays estimation results of non-parametric functions of GAMA associated with a  target size $\alpha = 0.05$ for the credit card dataset.
\end{tablenotes}
\end{threeparttable}
\label{tab:CreditCard_GAM_results}
\end{table}

\begin{figure}[!htbp]
	\centering
	\captionsetup[subfloat]{labelformat=empty,justification=centering}
	\subfloat[\qquad Random  Forest]{{\includegraphics[width=5cm]{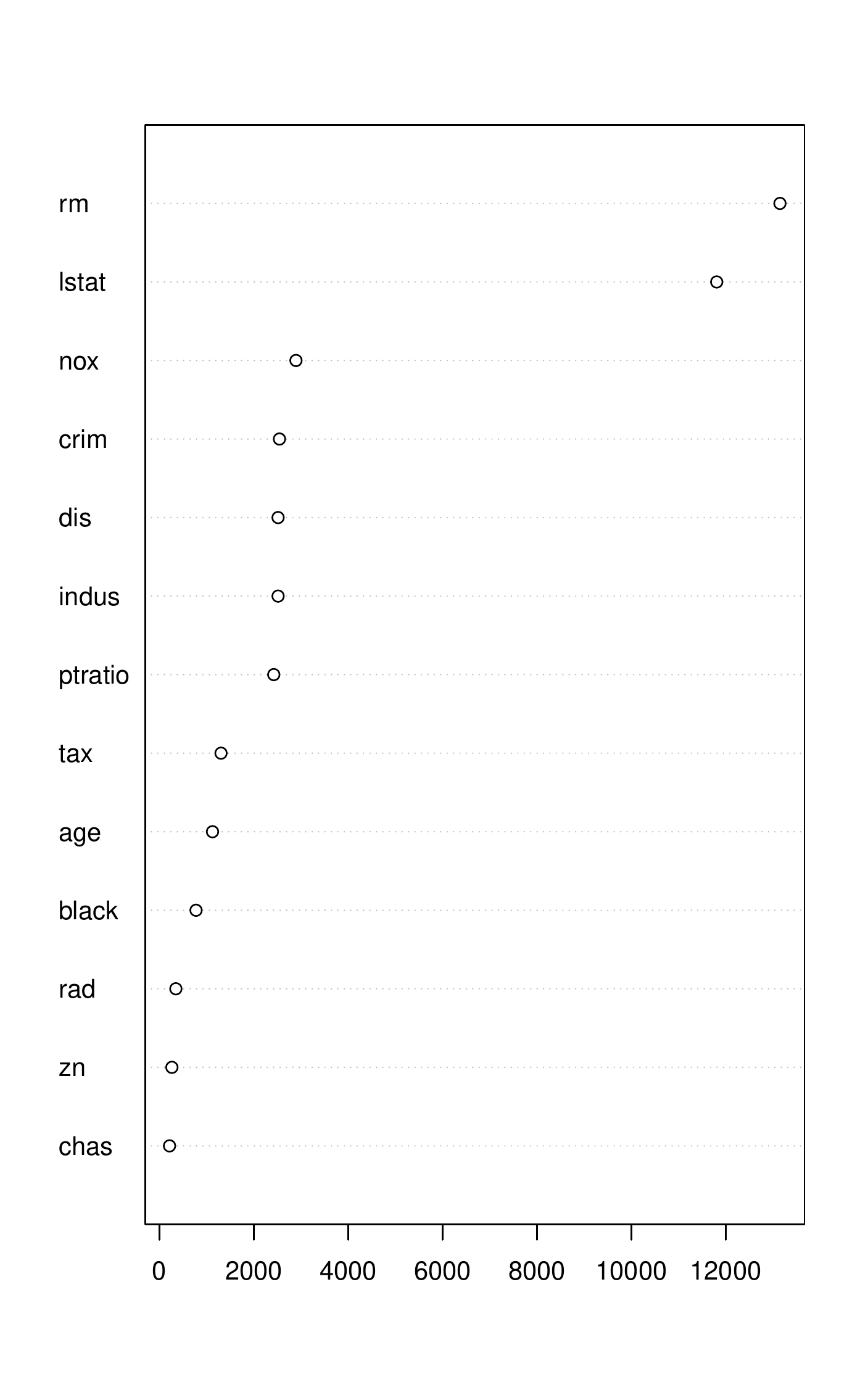} }}
	\qquad
	\subfloat[\qquad \qquad XGBoost]{{\includegraphics[width=5cm]{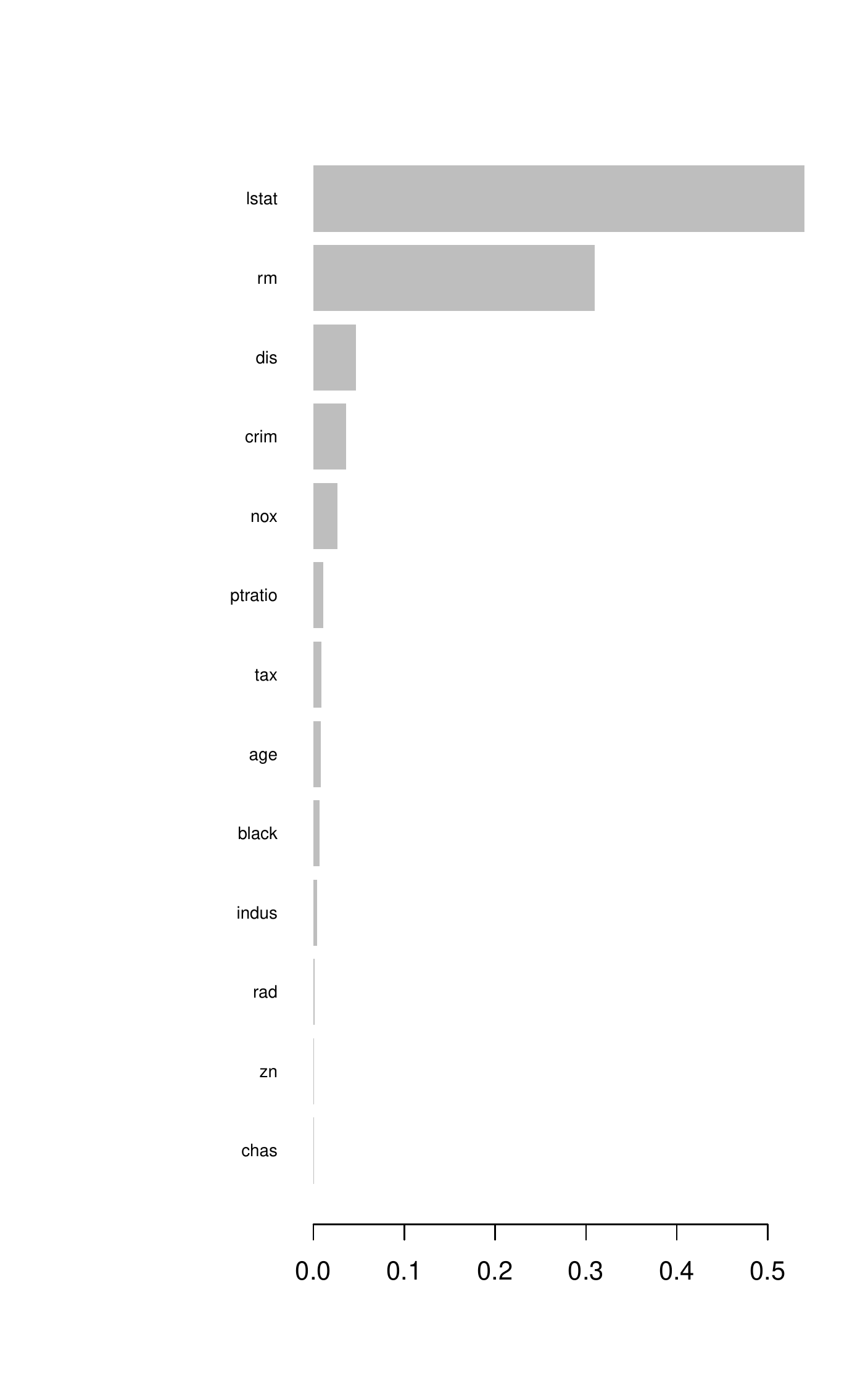} }}
	\caption{Variable importance of Random Forest and XGBoost algorithms}
	\label{fig:Variable_Importance}
\end{figure}

\newpage

\section{Penalized Logistic Tree Regression}\label{AppendixB}

Similarly to GAM(L)A, the Penalized Logistic Tree Regression (PLTR) model of \Citet{Dumitrescu2022} also aims at improving the predictive performance of traditional linear models by automatically capturing non-linearities. However, the method used to capture non-linear effects differs from that of GAM(L)A. To improve the predictive performance of the logistic regression in a classification problem's context, the PLTR is based on a two-step methodology relying on short-depth decision trees and a penalized regression. 

The objective of the first step is to identify threshold effects from decision trees with one and two splits. Firstly, a decision tree with one split is build for each explanatory variables leading to two leafs, independently of their level of information, which capture univariate threshold effects. As the two leafs are multicolinear, only the first one, denoted as $\mathcal{V}^{j}_{1}$, is retained to avoid multicollinearity issues. Secondly, a decision tree with two splits is build for each covariate couple leading to three binary variables. While the first binary variable accounts for univariate threshold effects, the second and third leafs capture bivariate threshold effects. Only one of the two latter leafs, denoted as $\mathcal{V}^{j,k}_{2}$,  is retained so as to account for two-splits threshold effects and avoid multicollinearity issues.

In a second step, these univariate and bivariate threshold effects are plugged in a logistic regression, such as
\begin{equation*}
	\Pr \left(y_i = 1|X,\mathcal{V}_{i,1},\mathcal{V}_{i,2};\Theta\right) = \frac{1}{1 + \exp\left[-\eta\left(X_i, \mathcal{V}_{i,1},\mathcal{V}_{i,2};\Theta\right)\right]},
\end{equation*}
where $\mathcal{V}_{i,1} = \left(\mathcal{V}^{1}_{i,1}, \dots, \mathcal{V}^{p}_{i,1}\right)$, $\mathcal{V}_{i,2} = \left(\mathcal{V}^{1,2}_{i,2}, \dots, \mathcal{V}^{p-1,p}_{i,2}\right)$,  
\begin{equation}
	\eta\left(X_i,\mathcal{V}_{i,1},\mathcal{V}_{i,2};\Theta\right) = X_i\beta + \mathcal{V}_{i,1}\xi + \mathcal{V}_{i,2}\zeta,
	\label{PLTR_eq}
\end{equation}
where
$\Theta = \left(\beta_0,\beta_1,\dots,\beta_p,\xi_1,\dots,\xi_p,\zeta_{1,2},\dots,\zeta_{p-1,p}\right)'$ is the set of $V$ parameters to estimate. The estimate $\hat{\Theta}$ is obtained by maximizing the following log-likelihood 
\begin{eqnarray*}
\mathcal{L}\left(y, X, \mathcal{V}_{1},\mathcal{V}_{2};\Theta\right) &=& \frac{1}{n}\sum_{i=1}^{N}\left\{y_i \log \left[F\left(\eta\left(X_i, \mathcal{V}_{i,1},\mathcal{V}_{i,2};\Theta\right)\right)\right] 
\right.  \\
&+& \left.
\left(1-y_i\right)\log \left[F\left(\eta\left(X_i,\mathcal{V}_{i,1},\mathcal{V}_{i,2};\Theta\right)\right)\right]\right\},
\end{eqnarray*}
where 
  $F\left(\eta\left(X_i,\mathcal{V}_{i,1},\mathcal{V}_{i,2};\Theta\right)\right)$ is the logistic cumulative density function. However, as the number of parameters to estimate can be relatively high, this approach is subject to overfitting issues. To solve this issue, the authors rely 
on the adaptive lasso estimator \citep{Zou2006}, defined as 
\begin{equation*}
	\hat{\Theta}_{alasso}\left(\lambda\right) = \argmin -\mathcal{L}\left(y, X, \mathcal{V}_{1},\mathcal{V}_{2};\Theta\right) + \lambda\sum_{v}^{V}w_v |\theta_v|,
\end{equation*}
where $w_v = |\hat{\theta}^{\left(0\right)}_v|^{-\nu}$, for $v=1,\dots,V$,  $\hat{\theta}^{\left(0\right)}_v$ is a consistent  estimate of  the $v$-th element of $\Theta$ and $\nu$ a positive constant. In their application, the authors rely on a logistic-ridge regression to obtain the $V$ parameters $\hat{\theta}^{\left(0\right)}_v$, set $\nu$ to 1, and $\lambda$ to $\hat{\lambda}^{min}$ via 10-fold cross-validation.

\end{document}